\documentclass[11pt]{book} 

\usepackage{tocloft}
\usepackage{pageslts}

\usepackage{ifxetex}

\ifxetex
	\usepackage{mathspec}
	\usepackage[no-math]{fontspec}
	\usepackage{xunicode}
	\usepackage{xltxtra}
	\AtBeginEnvironment{tabular}{\addfontfeatures{Numbers={Monospaced}}}
(Greek,Digits,Latin){Adobe Garamond Pro}
	\usepackage{etoolbox}
	\makeatletter
	\begingroup\lccode`~=`"
	\lowercase{\endgroup
	  \everymath{\let~\eu@active@quote}
	  \everydisplay{\let~\eu@active@quote}
	}
	\makeatother
\else
	 \usepackage[utf8]{inputenc}
\fi
\usepackage{textgreek}

\usepackage{moresize}

\usepackage{subfig}
\DeclareCaptionFont{ssmall}{\ssmall}
\DeclareCaptionFont{tiny}{\tiny}% "scriptsize" is defined by floatrow, "tiny" not
\usepackage{floatrow}
\DeclareFloatFont{tiny}{\tiny}% "scriptsize" is defined by floatrow, "tiny" not
\DeclareFloatFont{ssmall}{\ssmall}
\usepackage{tabularx}

\usepackage{url}

\addtocontents{toc}{\protect\setcounter{tocdepth}{1}}

\usepackage{emptypage} 

\usepackage{xspace}

\usepackage{blindtext}

\usepackage{graphicx}

\usepackage{booktabs}

\usepackage{longtable}
 
\usepackage{amsfonts,amsmath,amssymb}

\usepackage{pdfpages}

\usepackage{auxiliary_texfiles/aas_macros}

\usepackage[swedish,ngerman,british]{babel} 

\usepackage{natbib}

\usepackage[paperwidth=169mm,paperheight=239mm,total={13.3cm,19.6cm}, top=1.8cm, ignorehead, centering, footskip=\footskip+4mm ]{geometry}
\graphicspath{{Figures/}}

\newcommand\sect[1]{%
  \section*{#1}%
  \addcontentsline{toc}{section}{#1}
  \markright{#1}
}

\usepackage{fancyhdr}
\fancyheadoffset{0cm}
\usepackage{tcolorbox}
\tcbuselibrary{skins}
\newtcolorbox{thesisbox}[2][]{
oversize=-2.5cm,
  enhanced,
  attach boxed title to top left={yshift=-2ex,xshift=6ex},
  before skip=1em,
  after skip=1em,
  colframe=black,
  colback=white,
  fonttitle=\bfseries, 
  colbacktitle=white,
  coltitle=black,
  boxed title style={
    boxrule=0pt,
    colframe=white,
    },
  title=#2,
  #1}

\newcommand{\myName}{Zahra Gharaee}
\newcommand{\myYear}{2018}
\newcommand{\myMainTitle}{Action in Mind}
\newcommand{\mySubTitle}{A Neural Network Approach to Action Recognition and Segmentation}

\newcommand{\myFaculty}{Faculty of Humanities and Theology}
\newcommand{\myDepartment}{Department of Philosophy, Cognitive Science}
\newcommand{\myDegree}{Thesis for the degree of Ph.D.}
\newcommand{\myAdvisors}{Prof. Peter G\"ardenfors and Dr. Magnus Johnsson}
\newcommand{\myOpponent}{Prof. Angelo Cangelosi }

\newcommand{\myDefenceAnnouncement}{%
	To be presented, with the permission of the
	\myFaculty\xspace of Lund University, for public criticism at the \myDepartment\xspace at LUX, room C121 
	on Friday, the 27th of April \myYear\xspace at 10:00 am.}

\newcommand{\myAffiliation}{%
	\myFaculty\xspace of Lund University, \myDepartment\xspace.}

\newcommand{\myFundingInformation}{%
	The work was financially supported by Lund University, Cognitive Science and the EU FET project WYSIWYD.}

\begin{document}

%%%%%%%%%%%%% First pages with thesis title etc %%%%%%%%%%%%%%%%%%%%%%%%%%%%%%%%
% First six pages are contained in a separate file. Normally you should not
% need to edit it.
% This file contains the layout of the first six pages and the explanation text
% of a compilation thesis. 

%%%%%%%%%%%%% FIRST PAGES WITH THESIS TITLE ETC %%%%%%%%%%%%%%%%%%%%%%%%%%%%%%%%
\frontmatter % roman numbers

% Page one: Title only
%\thispagestyle{empty} % no page number
%\begin{center}
%\vspace*{5cm}
%{\Large \myMainTitle}
%%\\
%%{\large \mySubTitle}
%\end{center}

% Page two: empty. Empty space is important

%%%%%%%%%%%%%%%%%%%%%%%%%%%%%%%%%%%%%%%%%%%%%%%%%%%%%%%%%%%%%%%%%%%%%%%
% Page three: title page with small text (spikningsblad)

%\cleardoublepage
%\thispagestyle{empty} % no page number
%~
%\vfill
%\begin{center}
%{\HUGE \myMainTitle}
%\\[2mm]
%{\huge \mySubTitle}
%
%\vfill
%{by \myName}
%
%\vfill
%% black and white (default):
%\includegraphics[width=0.25\textwidth]{LundUniversity_C2line_BLACK.eps}
%
%\vspace{10mm}
%{\large \myDegree}\\
%{\large Thesis advisors: \myAdvisors}\\
%{\large Faculty opponent: \myOpponent}\\
%\vspace{1cm}
%{\footnotesize
%\myDefenceAnnouncement
%}
%\\
%\end{center}
%\vfill

%%%%%%%%%%%%%%%%%%%%%%%%%%%%%%%%%%%%%%%%%%%%%%%%%%%%%%%%%%%%%%%%%%%%%%%
% Page four: data sheet
%\newpage \thispagestyle{empty} % no page number
%%Either include datasheet texfile (will be filled in automatically), or a pdf
%%containing datasheet (text needs to be already in it, edit
%%sheetPDF_editable.pdf). Use one of the next two lines: 
%\input{auxiliary_texfiles/datasheet}
%%\addtocounter{pages}{1} 
%%\includepdf[width=1.40\textwidth]{datasheetPDF_editable_New}

%%%%%%%%%%%%%%%%%%%%%%%%%%%%%%%%%%%%%%%%%%%%%%%%%%%%%%%%%%%%%%%%%%%%%%%
% Page five: title and author, without small text. Looks good!

\cleardoublepage
\thispagestyle{empty} % no page number
~
\vfill
\begin{center}
{\HUGE \myMainTitle}
\\[2mm]
{\huge \mySubTitle}

\vfill
\vfill
\vfill
\vfill
\vfill

{ \LARGE \myName}

\vfill
\vfill
\vfill
\vfill
\vfill

{\Large
	\myYear
}
\vfill
\vfill
% black and white (default):
% \includegraphics[width=0.25\textwidth]{LundUniversity_C2line_BLACK.eps}

\vfill
\vfill
\vfill
{\large
	\myDegree
}

{\large
\myAffiliation
}
\vfill
\vfill
\vfill
{\normalsize
	\myFundingInformation
}

% Colour text in white so that the spacing is the same as on page three, but with less clutter
\color{white}{
\vspace{10mm}
{\large \myDegree}\\
{\large Thesis advisors: \myAdvisors}\\
{\large Faculty opponent: \myOpponent}\\
\vspace{1cm}
{\footnotesize
\myDefenceAnnouncement
}
}
\\
\end{center}
\vfill

\setcounter{page}{1} % Page Roman 1 of the frontmatter
\setcounter{tocdepth}{1}
\tableofcontents

% no page number on toc page:
\addtocontents{toc}{\protect\thispagestyle{empty}}

\newpage
\sect {Abstract}
Recognizing and categorizing human actions is an important task with applications in various fields such as human-robot interaction, video analysis, surveillance, video retrieval, health care system and entertainment industry.

This thesis presents a novel computational approach for human action recognition through different implementations of multi-layer architectures based on artificial neural networks. Each system level development is designed to solve different aspects of the action recognition problem including online real-time processing, action segmentation and the involvement of objects. The analysis of the experimental results are illustrated and described in six articles.

The proposed action recognition architecture of this thesis is composed of several processing layers including a preprocessing layer, an ordered vector representation layer and three layers of neural networks. It utilizes self-organizing neural networks such as Kohonen feature maps and growing grids as the main neural network layers. Thus the architecture presents a biological plausible approach with certain features such as topographic organization of the neurons, lateral interactions, semi-supervised learning and the ability to represent high dimensional input space in lower dimensional maps.

For each level of development the system is trained with the input data consisting of consecutive 3D body postures and tested with generalized input data that the system has never met before. The experimental results of different system level developments show that the system performs well with quite high accuracy for recognizing human actions.

\newpage
\sect {Introduction and the scope of the thesis}

As humans, we are extremely efficient in recognizing the actions of others. For example, we see immediately whether someone is walking or jogging, even if the movements are very similar. Furthermore, recognizing actions in the form of gestures or facial expression plays a major role in human communication. In modern society one finds ever more artificial systems that we are supposed to interact with in various ways. In particular, robotic systems, for example lawn mowers and vacuum cleaners, move around in our spaces. In the near future, we can expect many more such robotic systems that we have to interact with. Social robotics is a growing field with many different applications. Therefore an increasingly important scientific and engineering problem is to develop artificial systems that recognize or categorize actions in an efficient and reliable way. Solutions to this problem are important for many kinds of applications. 

One application example concerns interactions between a human and a robot assistant in health care situations. This scenario may be particularly relevant for old people who have some form of disability or memory condition. One can, for example, imagine that the robot helps the person to get in and out of bed, to open doors, to put out the trash, to help with laundry, and to assist in kitchen activities. To understand the activities and goals of the person, the robot must be able to attend to his/her motions and recognize the actions and the intentions behind them.

Another area where action recognition is important concerns surveillance systems. In this case, the system receives huge amounts of information about the scenes that are surveilled. This information may contain the physical movements of one or more individuals (humans or animals) acting in various ways depending on their different intentions. In such situations, an automatic action recognition system that can categorize the actions into meaningful classes would be very useful. Furthermore, if the system can interpret the intentions of the individual, it can predict possible future actions, which may prevent dangerous incidents or illegal activities.

A third area where action perception will be useful is within the entertainment industry, particularly in different forms of computer-based games where there is a high demand for human-computer interactions. In advanced games where the human body is involved, the gaming would become more advanced and challenging if the computer can identify the actions performed by the human. Again, if the computer can read the intentions of the individual, it can predict future actions and thereby make the game more strategically interesting.

There are many other applications for action recognition tasks. In general, to have an effective interaction with another agent like a robot, it is necessary, both for a human and for a robot, to perceive the motions and to recognize the actions of the other. Therefore, human-robot interaction is largely dependent on an efficient action recognition system. Advances on the action recognition problem will make the robot able to have richer interactions with humans. 

This dissertation is a contribution to this problem and its main purpose is to develop an efficient action recognition framework in a robotic context. My objective has been to develop systems, based on neural networks that can learn to recognize or categorize different human actions. I have also tested the systems using different types of input data – prerecorded and presegmented movies as well as online camera input where the system learns to categorize actions in real time. I have evaluated the systems in a number of experiments, in particular with respect to the accuracy of the categorization and the efficiency in learning a new set of action categories.

When developing the systems presented in this thesis, I have taken inspiration from how humans recognize and categorize actions. As humans we are equipped with a very efficient action recognition mechanism that we use automatically in our daily experiences. Experiments on human subjects reveal that they are capable of recognizing the actions performed by an actor after only about two hundred milliseconds of just observing the rough kinematics of the movements of the body joints (\cite{johansson}). Further experiments by \cite{Runesson1} and \cite{Runesson2} show that subjects extract subtle details of the actions performed, such as the gender of the person walking or the weight of objects lifted (where the objects themselves cannot be seen).

My aim has been to use available knowledge about human action recognition in implementing artificial systems. From a computational point of view, action recognition is quite difficult. First it requires a channel through which the artificial action recognizer communicates with humans. Humans and animals communicate with each other through sensory modalities such as vision, hearing, olfaction and touch. For example if you see your friend, then you wave your hand or nod your head to greet her and you do it only when you believe that she will see your action. The systems studied in this thesis use only the visual modality, when the system is interpreting human actions.

There are different types of visual input that can be used for action recognition systems, for example RGB images, depth maps and skeleton information. Each type has its advantages as well as limitations. Some of the action recognition methods are largely dependent on the types of input they utilize. For the aims of this research, I have provided the system with skeleton information of a moving human body collected by a Kinect camera. The input information consists of a vector of 3D positions of the joints of a human skeleton (see the cover of the book). In earlier research with an action recognition architecture that is similar to the neural network architecture of this study, 2D contours (black and white images) were also used as input (\cite{Buonamente1}).

In my studies, an action is represented to the action recognizer by a sequence of consecutive posture frames, where each frame is composed of 3D joint positions. Each posture represents the static pose of the action performer, while the consecutive posture frames show the kinematics of the action during a time interval (motions). This generates spatio-temporal characteristics of the actions, which is the central structure to be dealt with for the action recognition system.

If, for example, an action sequence is represented by on average 50 posture frames and each posture frame represents 20 skeleton joints in 3D space, then the system receives as input data 50 vectors of 60 dimensions just by observing a single action sequence that is performed. This illustrates that the problem involves a high-dimensional input space, which makes the perceptual analysis complicated. To deal with this condition, a method is required that maps from a high-dimensional input space to a low-dimensional space without omitting the substantial information of the action data. 

The multi-layer neural network architectures that have been designed and developed for this research are able to deal with different problems relating to action recognition tasks. The main modules of the proposed architectures are preprocessing, neural network layers and ordered vector representation. The preprocessing module modifies the input data to make the data independent of variation in viewpoint angles and change in distance and it employs cognitive functions like an attention mechanism and dynamics extraction to improve the performance of the system.    

The neural network layers perform several tasks in the architectures. First comes a mapping from the high-dimensional input space to 2D topographic maps that extract spatio-temporal features of the actions. Second, based on sequences of the extracted features, another neural map is formed with clusters or sub-clusters corresponding to different actions. Using self-organizing neural networks such as self-organizing maps or growing grids performs the first and the second tasks. The third task performed by the third-layer neural network is to label the clusters/sub-clusters of the previous neural map and thus categorize the actions.

For the aims of this study, I implemented different levels of action recognition architectures to improve the efficiency and robustness of the system and to make it function for a wide range of actions. First, I developed a system that recognizes the actions through the physical movements of the performer without involving any other entities. Next I developed a hybrid system that recognizes actions also involving objects, for example that the agent moves an object to a particular area. Then I developed the architecture to perform action recognition in an online real-time mode. Next, the action recognition system was developed to recognize unsegmented actions in online test experiments. Finally, a new architecture that builds on growing grid neural networks instead of the self-organizing maps was designed and implemented. This change in the underlying structure made it possible to perform the action recognition tasks more efficiently and the learning was much faster. 

In the following, I will, in Chapter 1, present the biological motion analysis as well as intentional actions that sometimes result in a change in the world as a consequence of the performance of an action. In Chapter 2, the action recognition problem and its challenges including the input space are described. Other proposed approaches of the literature for categorizing and recognizing human actions, which see the same problem from different perspectives such as human-robot interactions or by using language/concept for recognizing the actions, are presented in chapter 2. In Chapter 3, I describe my proposed action recognition approach together with the biological and cognitive inspirations used to develop different components of the system. The action segmentation problem is described in Chapter 4, together with psychological approaches to human action or event segmentation. Later in Chapter 4, I present the computational models for action segmentation. Next, In Chapter 5, I describe the experiments I have performed during my thesis work to study different angles of the action recognition problem. After these background chapters follow the six articles that form the core of this thesis.

% ===============================================================
% ======================= SUMMARY CHAPTER   =====================
% Back to British spelling
\selectlanguage{british}
% Need to add hyphenation corrections? Do like this:
% \hyphenation{as-tro-me-try ana-lysis}

% Page numbers arabic 
\mainmatter
% Reset table counters to not count the publications table
\setcounter{table}{0} 
% Rest page counters, this is where it all starts!
\setcounter{page}{1}

\chapter{Intentional actions vs biological motion}
\section{Introduction}

Before answering the question of how we perceive the actions performed by other humans, let us first consider how we produce an action. The muscles are activated by motor neurons that are the final neural elements of the motor system. The two prominent sub-cortical structures involved in the motor control are the cerebellum and the basal ganglia (\cite{Gazzaniga}).

The neural codes found in the motor areas are abstract and more related to the goals of the action than to the specific muscle patterns needed to produce the movement to achieve the goal. Thus the motor cortex may have more than one option for achieving a goal. As an example, consider the case when you are working on the computer, typing a letter, with a cup of coffee on your desk. Let us assume that you have two options: continue typing the letter or sipping your coffee. If you choose the coffee, then you need to achieve some intermediate goals such as reaching the cup, grasping it and bringing it to the mouth. Each of these intermediate goals requires a set of movements and in each case there is more than one way to perform them. For example, which hand do you chose to take the cup of coffee? In situations like this, you make your decisions on multiple levels and thus you choose a goal, choose an option to achieve the goal and also choose how to perform each intermediate step.

The affordance competition hypothesis proposed by \cite{Cisek} builds on an evolutionary perspective, which says that the brain's functional architecture  has evolved to mediate real-time interactions with the world. These interactions are driven by the internal needs of humans, such as thirst and hunger, while the world defines the opportunities for the actions, the so-called affordances.    
   
The affordance competition hypothesis claims that humans develop multiple plans in parallel. While performing an action we are already preparing the next steps. This means that both the process of action selection (what action to chose) and its specification (how to perform the action) take place simultaneously within an interactive neuronal system.

 \section{Event perception in humans}
I next turn to how humans perceive events. There are three main theories in this area: the perception of causality, the ecological approach and biological motion. Firstly, causality is an important aspect of the events and it plays an important role in event cognition. In a famous series of experiments, \cite{Michotte} investigated the role of causality in the perception of events. He studied how the observer perceives the event when viewing animated sequences involving a small number of objects. In one class of experiments, a square moved in a straight line from left to right until it approached a second square. Then the first square stopped moving and the second started to move in the same straight pathway as the first one.

By changing the parameters such as the absolute and the relative speed of the objects, the distance between them when the first object stops and also the time between when first object stops moving and the second starts to move, Michotte identified a range of parameters that led to the perception that the first object causes the movement of the second object. The perception of causation was strongest when (1) the movement of second object starts at the same time that the movement of the first object ends, (2) the two objects were not too far apart, (3) they moved on the same motion path but not too slowly and (4) the second object moved at a slower speed than the first one.

Michotte argued that the critical determinant in perceiving the causal interaction between two objects is when the motion of the objects is perceived as a single event. He called the perception of common motion across different objects the \textit{ampilation} of the movement. In a case of launching, the motion of the first object is transferred to the second one when the first stops and the second starts moving. In the condition that the time interval between the stopping of the first one and the starting of the second one is long, the observer perceives two separate motion events with no causal interaction. 

A general problem regarding the ampilation of the movement is which computational principles that govern it. Michotte argues that the ampilation is determined by a wide set of principles related to Gestalt continuity laws. These principles don't depend on experience with particular interactions but they are a priori aspects of the human perception structure. Partly because of the measurement issue, no complete account of ampilation of the movement and its connection to the perception of causality has been proposed, which makes this theory difficult to evaluate. Other researchers who replicated the Michotte experiments also reported that the causal judgments are sensitive to learning, expectations and context.

Secondly, in the ecological approach to perception, \cite{Gibson2} considers three main kinds of events in visual perception. The first concerns changes in the layout of surfaces, the second changes in the color or texture of the surface, and the third the coming into or out of the existing surfaces. As an example, assume that someone puts a ball into a glass of milk. If the ball density is a little less than that of the milk, then the ball barely floats in the glass. This leads to changes in the surface layout and to the creation or destruction of surfaces.

Gibson argues that an event is determined by the presence of an invariant structure that persists throughout the changes. This approach relies on the structure of the world surrounding the observer, while it does not rely on the experience or mental structure of the observer. Gibson’s approach may account for simple visual events determined through physical properties. It can not, however, describe mental representations of events and the role of these representations in cognition.

The third approach is biological motion as investigated by \cite{johansson}. It concerns the perception of a moving body when a \textit{point-light} technique is applied. This approach will be described in more detail in the following section \ref{Motion Analysis}. Johansson’s \textit{point-light} experiments played an important role for a theory of how actions are individuated, identified and represented. It inspired much later research on biological motion analysis.

The three types of research on the perception of actions and events presented here indicate two common conclusions. The first is that the dynamic features of activities are central, which means that an event can be perceived with a trajectory of changes over time (spatio-temporal trajectory). The three theories all agree that what individuates the event is a configuration that is present over the time duration in which the event lasts. Therefore, although the events consist of changes, they still possess a higher-order stability that persists through these changes, and this stability is what characterizes them. The second conclusion is that the perceptual systems organize the information in a hierarchical manner. The sensory components are combined into more general forms in this hierarchy.

In the real world, however action perception occurs in multiple ways in humans. For example we can recognize some actions without dynamics like the action ‘point’. In fact the recognition often only takes one body posture, or one body posture in a particular context. With that said, in this thesis I will study the actions based on their spatio-temporal trajectories.
     
\section{Motion analysis} 
\label{Motion Analysis}
The term biological motion is used for any type of physical movements that are performed by humans or other animals. Biological motion is almost always meaningful for both the performer and the observer. For human actions and for many biological motions, we categorize them with the aid of verbs in language, for example, \textit{wave}, \textit{walk} and \textit{punch}. Some biological motions are pure movements such as the movements that carried out in dance or sports. We can still describe them partly by using verbs such as move arm up/down, move leg back/forward, turn head right/left etc.

For his studies, \cite{johansson} designed experiments based on a patch light technique for visual perception of motion patterns characteristic of living organisms like humans. In this technique, the actions are represented by a few bright spots describing the motions of the main joints of a human while an action like walking, running, dancing, etc., is performed. 

The patch light technique is built up from, bright (or dark) spots moving against a homogeneous contrasting background. The kinematic-geometric model for visual vector analysis that was originally developed for studying perception of mechanical motion patterns was extended to biological motion patterns.

An example of the mechanical perception studies is Wertheimer’s demonstration of the “Law of common fate”. When, in a previously static pattern of dots, some dots began to move in a unitary way, the static form is broken up and the moving dots form a new unit. In an analogous way, the joints of a human body are the end points of bones with constant length and their motion is seen as the end points on a moving, invisible rigid line or rod. Johansson calls this perceptual mechanism ‘the rigidity principle’.

To generate the stimuli for the perception of biological motion a video of an actor, dressed in black against a black background, was recorded. Small patches of reflecting tape were attached to certain areas of the actor's body, representing the actor's joints and they were flooded by light from one or two searchlights mounted close to the lens of the video camera. The recording begins when the actor starts performing the action and lasts to the end of the action. The movie of the motions of $10$ spots from the main joints of a human body has always given the impression of a human performing the action (for example walking in frontal direction). It is important to point out that when the motion was stopped, the set of light spots was never interpreted as representing a human body by the observers – it is only the dynamic pattern that generates a perception of an action.

\begin{figure}
\includegraphics[width=0.5\textwidth]{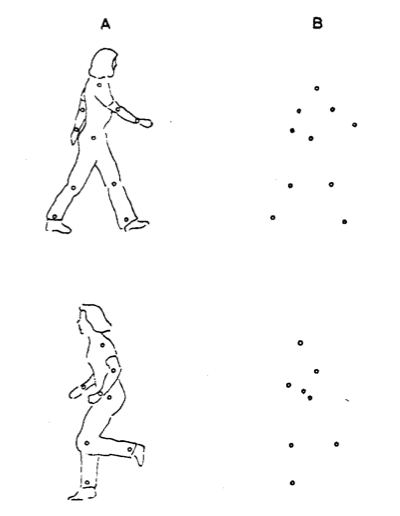}
\caption{The contours of a walking (above) and running (below) subject shown in part (A). The corresponding dot configuration is shown in part (B). From \cite{johansson}.\label{fig:fig0}}
\end{figure}

Now, the question is how $10$ points moving simultaneously on a screen in a rather irregular way can create such a vivid and definite impression of human walking or jogging (see Fig.~\ref{fig:fig0}). In the research by \cite{johansson}, it was the first time ever for the observers to see a walking pattern built up from moving light spots. So the question is why these spots evoke the same impression of motion as a movie of a walking person does. 

A first answer is that the previous experiences of the observers help them to recognize the human walking. There exists a heavy over-learning in seeing humans walking, which makes it natural for us to perceive the motion pattern of the light dots as a walking human. The first guess could be that the grouping of moving elements is determined by general perceptual principles, but the vividness of the perception is a consequence of prior learning.

\begin{figure}
\includegraphics[width=0.6\textwidth]{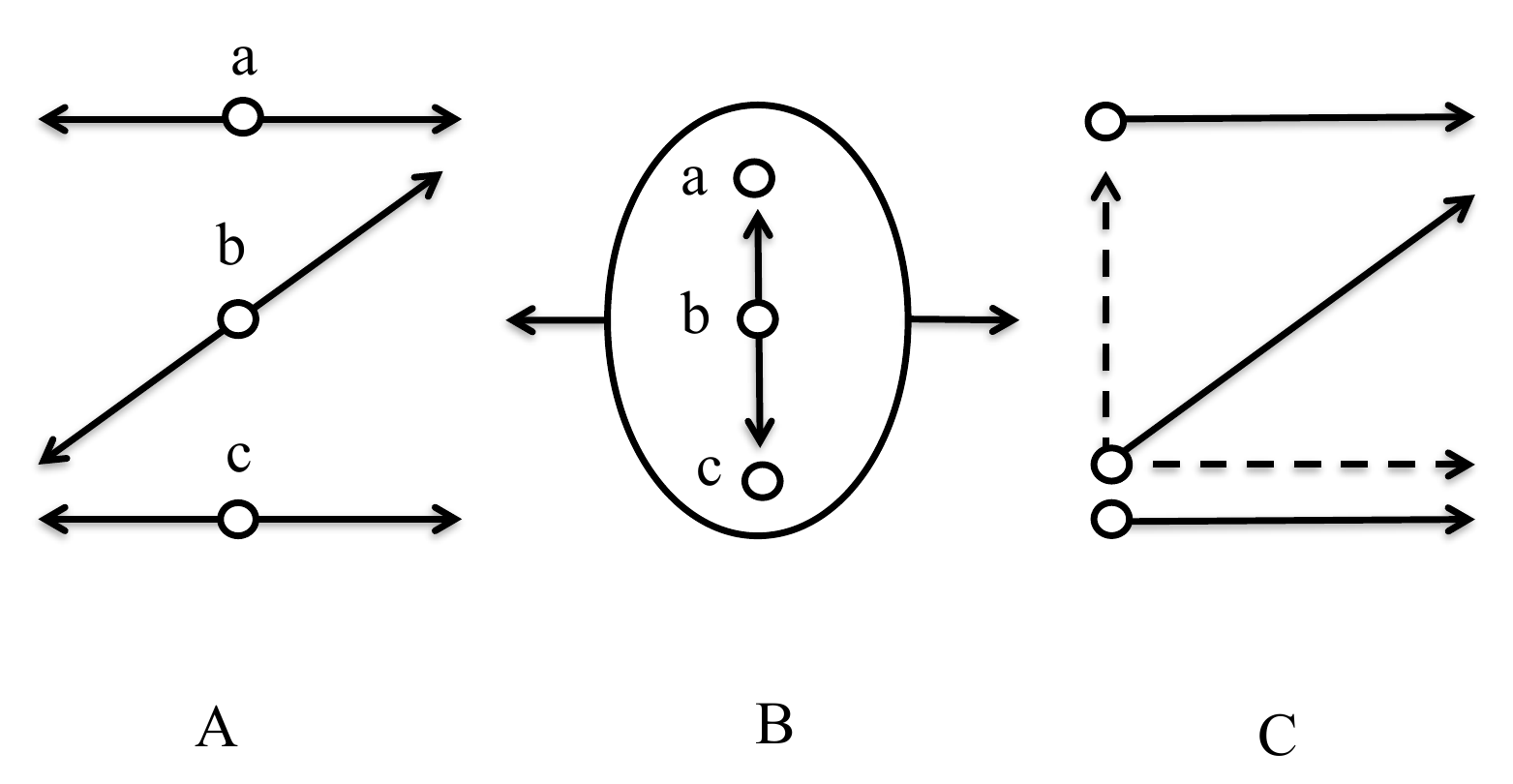}
\caption{An example of visual vector analysis. The proximal pattern of the moving dots (A). The perceived diagram extracted from the stimuli combination (B). The resulting vector analysis of the motion of the middle point corresponding to the perception.\label{fig:fig1}}
\end{figure}

To find a more complete answer, it is useful to briefly consider visual perceptual processes and to understand how the vision system constructs the perception. The model for motion and space perception is called visual vector analysis and it has three main principles: First, the elements in motion on the picture plane of the eye are always perceptually related to each other. Second, the equal and simultaneous motions in a set of proximal elements automatically connect these elements to rigid perceptual units (following Johansson’s rigidity principle). Finally, the third principle says that when in the motions of a set of proximal elements, equal simultaneous motion vectors can be mathematically abstracted these components are perceptually isolated and perceived as one unitary motion. 

\begin{figure}
\includegraphics[width=0.5\textwidth]{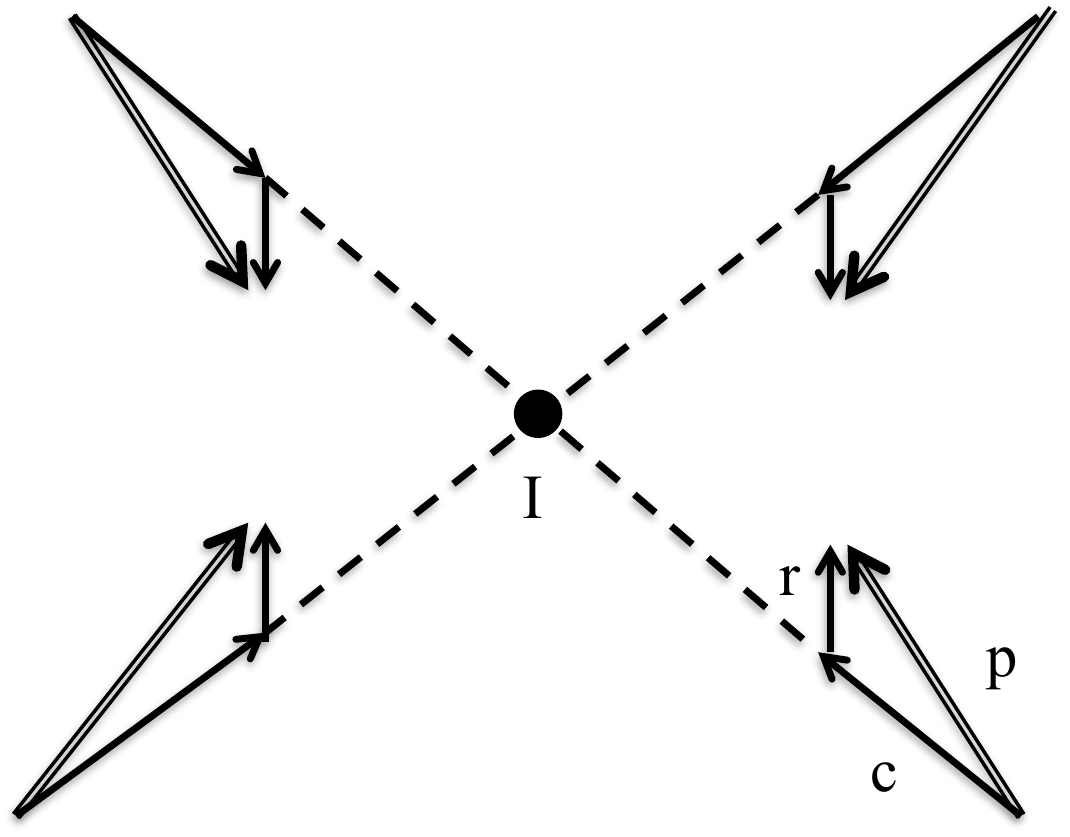}
\caption{Vector analysis of a non-respective shrinking configuration represented by four elements. The physical motion of the element (p). The common concurrent motion component (C) is perceived as a translation to the depth. The residual (r) is perceived as a shrinking or as a rotation in depth.\label{fig:fig2}}
\end{figure}

The term equal in the second and the third principles does not only refer to the Euclidean parallel dot motions with the same velocity but also includes, first, the motions that follow tracks that converge to a common point in depth (a point at infinity) on the picture plane and, second, dot motions where their velocities are mutually proportional relative to this point. 

The visual vector analysis specifies the basic mechanism for visual motion and space perception. It has the consequence that the ever-changing stimulus patterns on the retina are analyzed in order to detect maximal rigidity in coherent structures. The mechanism explains why we automatically obtain perceived size constancy as well as form constancy from projection of rigid objects in motion. What is critical for perceiving rigidity is not rigidity in distal objects, but rather the occurrence of equal motion components in the proximal stimulus. This process is fully automatic and independent of cognitive control.

To understand the visual vector analysis better let's assume that there are three moving points named as \textit{a}, \textit{b} and \textit{c} as shown in Fig. ~\ref{fig:fig1} (part A) in which points \textit{a} and \textit{c} are moving horizontally and point \textit{b} is moving in a diagonal direction. The observer perceives the pattern as three dots forming a vertical line moving horizontally while point \textit{b} also moves vertically up and down along the line as shown in Fig.~\ref{fig:fig1} (part B). The vector analysis for the motion of the point \textit{b} is shown in Fig.~\ref{fig:fig1} (part C) in accordance with the perception description. The diagram demonstrates the correctness of the analysis from a mathematical point of view.

\begin{figure}
\includegraphics[width=0.6\textwidth]{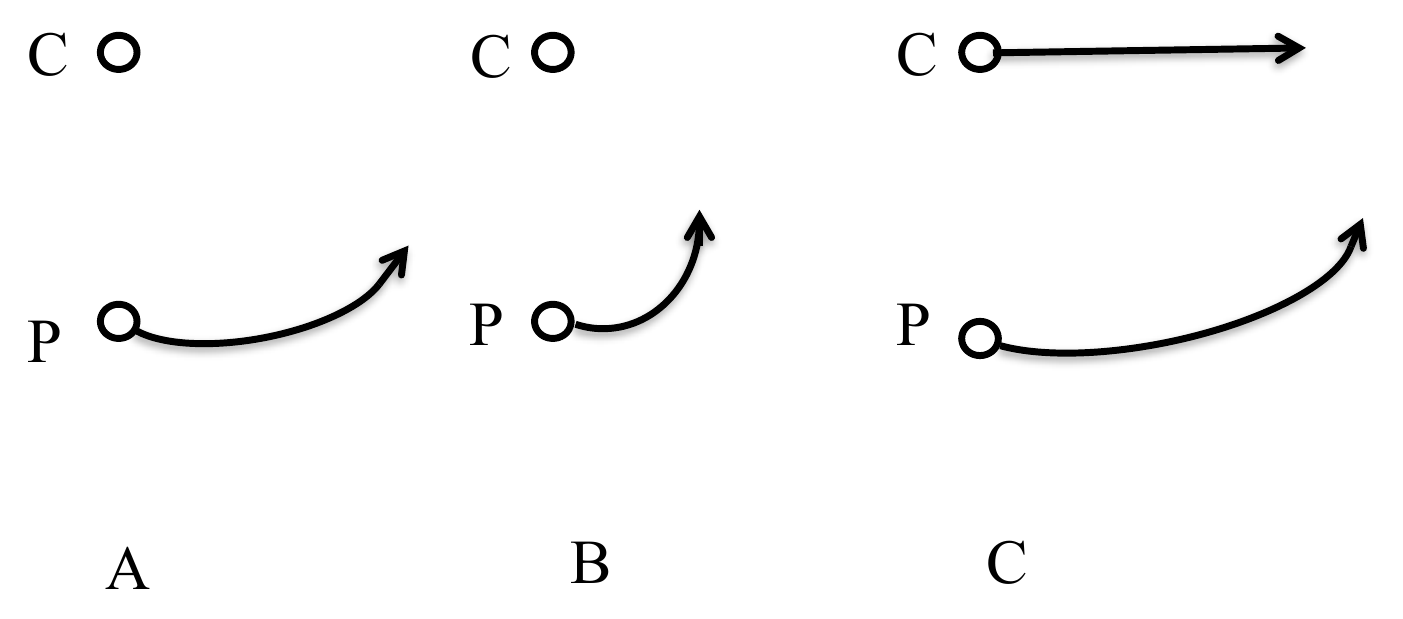}
\caption{Three motion combinations of a pendulum in two stimuli. The points are perceived as a fronto-parallel pendulum motion of point P about point C while there is a constant distance between the two points but change of the angle (A). The points are perceived as a pendulum motion in plane in a certain angle (30-60 deg) towards the fronto-parallel plane (both distance and angle change) (B). The same motion as perceived in (B), but the axis is now moving during the cycle (C).\label{fig:fig3}}
\end{figure}

Another illustrative example of visual vector analysis is shown in Fig.~\ref{fig:fig2}. Consider four elements all moving together toward direction \textit{P}. The motion that an observer perceives is a shrinking and rotation in depth in relation to the point at infinity \textit{I}. The mathematical abstraction explains the motion by dividing the vector \textit{P} into two vectors \textit{r} and \textit{c}, where the vector \textit{r} represents the circular motion and the vector \textit{c} represents the translator motion in depth.

As a third example, take two elements perceptually representing the end points of a rod in a pendulum motion shown in Fig.~\ref{fig:fig3}. Experiments have shown that changing the angle, but keeping a constant distance between the two elements results in perceiving a frontal-parallel pendulum motion that is depicted in Fig.~\ref{fig:fig3} (part A). Changing both the angle and the distance between two elements results in perceiving a pendulum motion in depth, which is shown in Fig.~\ref{fig:fig3} (part B). Now if a constant component of translation is added to both elements of the motion pattern, as shown in Fig.~\ref{fig:fig3} (part C), the motion will be seen as the same as the pendulum motion in Fig.~\ref{fig:fig3} (part B). This means that the constant component is effectively separated perceptually and seen as a reference frame for the pendulum motion.

Such a perceptual separation of a common component is a typical example of perceptual vector analysis. When the common component is subtracted, the residual motion forms a translatory, rotary or pendulum motion. The common motion is just a reference frame for the deviating component and changes in this component do not change the perceived primary motion.

The patterns of biological motion can be described by subtracting the common motion components such as the semi-translatory motion of the hip and shoulder elements (the trunk), which are found in the movements of all elements in the body. In contrast, the motions of the knees and the elbows are rigid pendulum motions relative to this reference frame. The semi-translatory motion in walking, which is inherent in the movements of all $10$ elements, plays no decisive role for identifying a walking pattern.

To investigate whether the recognition of walking from the motion of $10$ points is independent of the course of the common component, three experiments were performed by \cite{johansson}. In the first experiment, a common component was subtracted from the element movements in a walking pattern. The horizontal translatory component was subtracted and the up-and-down motion remained as a small common motion residual. The result shows that the observers still immediately reported seeing a walking person.

In the second experiment, the amount of information given to the observer was manipulated so that the observer was shown a little less than half a step cycle and this interval was chosen randomly during the walking period. The result of the second experiment also shows that all observers without any hesitation reported a walking human.
 
Finally, in the third experiment, in contrast to the first two experiments, an extra component was added to the primary movement of each element. The extra component was produced by slowly rotating a mirror oriented at a $45$-degree angle to the optical axis of the lens in front of the camera. The mirror was reflected to the TV camera by which the scene was recorded. The circular motion produced had a diameter of a circle covering about one-third of the distance between the foot and the shoulder element. The result of the third experiment shows that all subjects also reported seeing a walking man, albeit in a very strange \textit{wavy} way.
 
The general conclusion of these three experiments is that adding or subtracting common components to the movements of the elements does not disturb the identification of the walking pattern. It also shows that the vector analysis model is valid for the perception of complex motion patterns representing biological motions.

The fact that the natural kinematic patterns, based on the point-light displays used by \cite{johansson}, contain unique information about the dynamic properties of the elements is called Kinematic Specification of Dynamics (KSD) by \cite{Runesson1}. In the experiments performed by \cite{johansson}, they applied the KSD information by using the recordings of natural event. Normally visible features such as shapes, colors, clothing and facial expressions were removed as well. Therefore only rough kinematic aspects were available to the observers.

The perceptual outcome of the KSD is much richer than only representing that the persons engaged in walking or other activities are detected and recognized from the recordings by the human observers. For instance, the observer can perceive how strenuous push-ups are and that a person who climbs up on a ladder does not start painting a wall and only pretends to do so. They can see that a bicycle ride is not real – it is indeed a pan-shot of a ride on a stationary exercise bike.

Hence, it appears that the information about inner causal factors behind an action is also conveyed by the point-light displays. In fact, there is efficient information concerning a number of finer properties of the person and the actions in patch-light displays, like the mass of an object handled by a person, the expectations or deceptive intentions they might have concerning their mass and the gender or identity of the person (see \cite{Runesson1} and \cite{Runesson2}).

In fact subjects perceive the weight of an object handled by a person in the patch light experiments, even if the object itself is not visible in the movies. The main reason they can do it is exactly the same as for seeing the size and shape of a person’s nose or the color of the shirt in normal illumination, namely that the information about all these properties is available in the optic array. The only thing is that the kinematic-optic properties that specify the weight of a box may be analytically much more complex than the ones for the size of the nose. Therefore it is unwise to assume that the type of information such as the color of the shirt is fundamentally simpler or more accessible than the type for the weight of a lifted object (see \cite{Runesson1} and \cite{Runesson2}). 

Using the KSD approach helps us understand the sensory processing of biological motion. Using vision and applying visual vector analysis leads us to identify and extract the particular motion patterns such as the pendulum motions of the ankles or knees, the rotatory motions of the wrist, translatory motions of the shoulders and hips, etc. On the next level we then apply higher cognitive mechanisms to categorize them and to relate the motion patterns to different goals.

In other words, throughout life we gradually build a hierarchical perceptual mechanism in order to first perceive the biological motions on a primary sensory level and then relate them to the actions based on the goals they follow on higher processing levels. As an example, when the pendulum motion patterns of the knee and the ankles together with the rotatory motion patterns of the wrists are identified by the principles of visual vector analysis, we use our concepts and prior knowledge to categorize these detected motion patterns as \textit{walking}. By observing various types of walking humans, our conceptual space will become richer so that recognizing a walking person will happen quicker and easier with time.

Johansson's patch-light technique has been used in later research, for example by \cite{Giese2}. In order to study how actions are categorized, they recorded subjects performing actions like walking, running, limping and marching. After that they generated morphs of the actions recorded by creating linear combinations of the dot positions that appeared in the recorded videos. The morphed movies thus show actions that are mixtures of the original ones. Then they asked the observers to categorize the morphed videos into walking, running, limping or marching as well as to judge the naturalness of the actions. The result shows that there are clear prototype effects in the categorizations (for another study of prototype features of action categorization see \cite{Hemeren}).

In another study presented in \cite{Giese1}, it was shown that the visual representations of complex body movements can be characterized as perceptual spaces with metric properties. They develop a measure of the similarity between the perceptual metrics and the metrics of a physical space that is defined by distance measures between joint trajectories. They construct a mapping between physical and perceptual spaces that preserves distance ranks (second order isomorphism), which provides a representation useful for the classification and categorization of motion patterns based on their physical similarities. 

To test the measure, \cite{Giese1} performed two experiments and one control experiment. Parameterized classes of motion patterns were created by motion morphing, applying a method that generates new movements by linear combination of trajectories of three prototypical gait patterns (walking, running and marching). The perceived similarities of these patterns were assessed using two different experimental paradigms. Based on the perceived similarities of pairs of the motion patterns, the metrics of the perceptual space were reconstructed by multidimensional scaling (MDS) and the recovered configurations in perceptual space were compared to the original configuration in morphing weight space.

  \section{The role of the results of actions in event descriptions}
So far, I have considered the motion patterns as the only source of information when categorizing an action. In this section, I will investigate the roles of the objects that may be involved in actions. 

There are two main ways of describing an event. The first way focuses on motion patterns, as described above, and thereby the manner of the action. The second way to describe events focuses on the result of the action. With the second approach, other sources of information than motion patterns are utilized in event description. This new information contains changes of the objects or other entities involved in the performance of the actions. 

To have a better understanding of the two approaches for recognizing the actions, first we should have a closer look at different types of actions. In the action world there are ones that only express a particular manner without any involvement from other entities such as objects. For example in the action \textit{wave}, based on how it is performed, there are rotatory and translatory motions of the wrists and arms, which are performed to satisfy a goal such as greeting or attracting someone's attention. No other entity is involved in completing the action \textit{wave} and as a consequence nothing is changed in the environment by doing this action.

On the other hand take the action \textit{push} performed by using a hand. In this case, based on how the action is performed, there is a translatory motion of the hand that applies the force in a particular direction to another entity such as a \textit{cup}. Here performing the action \textit{push} is completed only by having another entity that is involved in it. Therefore there will be a change in the environment as a consequence of doing the action \textit{push}, such as a move in a \textit{cup}.

Here I distinguish both the groups of actions described earlier. The first group could be named manner actions. They only express a manner without a particular detectable resulting state in the environment. The second group is named result actions, which are identified by a change of an object that is represented as a resulting state.

In other words, to understand an event, we usually build links between intentions, actions and consequences. Events can be described in terms of their causes and effects. Causes and effects are understood in terms of transfer or exchange of physical quantities in the world, such as energy, momentum, impact forces, chemical and electrical forces (\cite{Lallee}). There is also nonphysical causation, such as forcing someone to do something or to make a decision. This mental type of causation is understood by analogy with the physical forces. The causes involved in an event are typically described by manner verbs, while the effects are described by result verbs (\cite{Warglien} and \cite{Mealier}).

Although the dynamic forces are not directly perceivable, our visual experiences are very efficient in extracting the shape, position, direction, velocity and acceleration of a moving object. This means that our perceptual mechanism can immediately extract the forces involved in an action in accordance with Runesson’s KSD principle. As humans we normally utilize our vision as a resource of information to interpret the causal relationships between entities in the world, but auditory or haptic information can also be a complement.  

In another study presented in \cite{Marocco}, experiments are performed where actions result in changes in an object or other entities. In their study a simulated iCub humanoid robot learns an embodied representation of actions through the interaction with the environment as well as linking the effects of its own actions with the properties of the object before and after the action.  

Experiments testing recognition of actions involving an object are presented by \cite{Gharaee2, Gharaee4}. In the experiments illustrated by \cite{Gharaee4} a hybrid hierarchical architecture is implemented, which is capable of recognizing the actions performed as well as identifying the object involved in the action performance.

\chapter{Action recognition}
\section{Introduction}
Motion perception and action recognition are cognitive processes that play important roles in different aspects of the lives of humans and other animals including communication, mind reading, prediction and intention reading of others. By perceiving the actions of others we usually interpret them as intentional and we use this knowledge for a better understanding while interacting with others. For example, an arm wave could be interpreted as a greeting whereas pointing to a person, thing or location could attract attention towards another entity. 

Recognizing an action is necessary in order to select a reaction that is suitable for the specific condition. For instance, consider that one has an appointment for a job interview with the boss of a company. The applicant arrives at the office and the boss extends his hand to greet the applicant. What if the applicant does not recognize the action and the intention behind it or what if the applicant shows no reaction or even worse he/she reacts in the wrong way (for example, scratches his/her head)? 

Our perception of the actions performed by others can even be crucial for our survival. For example, encountering dangerous behavior from a person or an animal requires an immediate and suitable reaction. There are many other examples that show the importance of action recognition and categorization for humans and other living organisms. 

In this chapter, I will investigate action recognition in a more technical and systematic perspective. The first question is why we need artificial action recognition systems. There are indeed several applications for these systems including video surveillance, human-computer interaction, video retrieval, sign language recognition, robotics (social robotics), health care, video analysis (for example, sports video analysis) and computer games.

\section{Input for the action recognition systems}
The data representing the peripheral input space of the action performed is an important part of an action recognition system. Another question is what types of data should be used as the input to the system. This will be determined by the sensors utilized to collect the action dataset. For any types of sensors, the data consists of movies representing actions. The movies of actions can have different lengths since different actions can be performed during different time intervals.

Traditional action recognition methods use consecutive sequences of images. Recognizing actions from image sequences taken by ordinary cameras has limitations. They are, for example, sensitive to color and illumination changes, occlusions, and background clutters. 

Another method is to use range sensors, However, earlier types of sensors were either too expensive, provided poor estimations, or were difficult to use on human subjects. For example, sonar sensors have a poor angular resolution and are susceptible to false echoes and reflections. Infrared and laser range finders can only provide measurements from one point in the scene. Radar systems are considerably more expensive and typically have high power consumption requirements. Motion capture by such sensors is expensive and more difficult for data collection.

The advent of the cost-effective RGB-D sensors such as $Microsoft Kinect^{TM}$ and $Asus Xtion^{TM}$ added another dimension, the depth, which is insensitive to illumination changes and provide us with the 3D structural information of the viewed scene. Moreover, the depth cameras can work in total darkness. This is a benefit for applications such as patient/animal monitoring systems, which run 24/7. The new motion analysis methods based on the RGB-D data are important consequences of this development. RGB-D data for human motion analysis provide us with three main types of information: RGB and depth and skeleton.

The input types adopted for an action recognition task play an important role in developing efficient methods. Some methods function well with specific types of inputs, but cannot always be generalized to all of the available types. Using space-time volumes, spatio-temporal features and trajectories, human action recognition tasks have been performed using RGB images (color images). For instance, \cite{Schuldt} proposed a method that extracts the spatio-temporal interest points and couples it with a support vector machine to recognize actions. Cuboid descriptors were utilized by \cite{Dollar} and SIFT feature trajectories modeled in a multi-layer system proposed by \cite{Sun}. Other methods to extract the spatio-temporal features from color images for recognizing human actions are proposed by \cite{Laptev2}, \cite{Bobick} and \cite{Davis}. 

The main characteristic of RGB data is the shape, color and texture information they provide that helps extract points of interest and optical flow. On the other hand, depth data is insensitive to variations of illumination, color and texture and it provides 3D structural information of the scene. 
  
With the advent of RGB-D sensors, action recognition methods were developed based on the depth maps, which mainly work by extracting spatio-temporal features. In holistic approaches, the global features such as silhouettes and space-time information are extracted. Such a methodology is utilized in several research articles such as \cite{Oreifej}, \cite{Rahmani}, \cite{Li}, \cite{Vieira},\cite{Yang-Xiaodong3} and \cite{Yang-Xiaodong2}. Other approaches extract the local features as a set of interest points from depth sequences (spatio-temporal features) and compute a feature descriptor for each interest point (for more see the methods proposed by \cite{Laptev1}, \cite{Wang1}, \cite{Wang2}, \cite{Ghodrati}, \cite{Xia2} and \cite{Wang-Jiang}). 
 
The skeleton data obtained from, for example a Kinect sensor, is more robust to scale and illumination changes and can be invariant of the camera point of view as well as body rotation and the speed of motion. The cost-effective depth sensors are then coupled with the real-time 3D skeleton estimation algorithm introduced by \cite{Shotton}. Most of the skeleton-based methods utilize either the 3D locations or the angles of the joints to represent the human skeleton. By extracting the spatial-temporal features from the 3D skeleton information, such as the relative geometric velocity between body parts, relative joint positions and joint angles in \cite{Yao}, the position differences of the skeleton joints in \cite{Yang-Xiaodong1} or the pose information together with differential quantities (speed and acceleration) in \cite{Zanfir}, the body skeleton information in space and time is first described. Then the descriptors are coupled with Principle Component Analysis (PCA) or some other classifier to categorize the actions. There are other methods in the literature using skeleton data for human action recognition such as \cite{Chaudhry}, \cite{Wang-Chunyu}, \cite{Miranda}, \cite{Vemulapalli} and \cite{Eweiwi}

At the same time, using the estimated 3D joint positions for human action recognition is limited. For example, the 3D joint positions are noisy and may have significant errors when there are occlusions such as one leg being in front of the other, a hand touching another body part, two hands crossing, etc. The estimation is not always reliable and can fail when the person touches the background or when the person is not in an upright position (e.g. a patient lying on a bed). Moreover the 3D skeleton motion alone is not sufficient to distinguish some actions. For example \textit{drink} and \textit{eat} generate very similar motion patterns for a human skeleton. Extra input, such as information about the objects involved, needs to be included and exploited for better recognition of the action.  

In some other methods, a fusion-based feature for the action recognition is applied, for example the method proposed by \cite{Zhu} in which the spatio-temporal features and the skeleton joints are fused as complementary features to recognize human actions. Another method that uses multi-fused features to recognize human actions is the Human Activity Recognition (HAR) system proposed by \cite{Jalal}. This method fuses four skeleton joint features together with one body shape feature representing the projections of the depth differential silhouettes between two consecutive frames onto three orthogonal planes.       

There are neural-network-based approaches developed to solve the problem of action recognition such as the methods developed by using the Convolutional Neural Networks (CNN) and the ones based on the Recurrent Neural Networks (RNN). The CNN-based models have had great success in dealing with the image-based tasks (see \cite{Karpathy} and \cite{Ng}) and the RNN-based methods are mainly suggested for the sequence-based tasks (see \cite{Liu}). Among skeleton-based motion recognition approaches with deep learning are the CNN-based methods proposed by \cite{Hou}, \cite{Wang-Pichao} and \cite{Liu} and the RNN-based approaches proposed by \cite{Du1}, \cite{Du2},\cite{Veeriah} and \cite{Zhu-Wentao}. 

There are, however, major challenges in running action recognition experiments. Here I present three major challenges of the vision-based human action recognition. The first is intra-class variability and inter-class similarity of the actions. In real-life recordings, the individuals perform one type of action in different directions with different characteristics of body part movements. Furthermore, two different actions may only be distinguishable by using very subtle spatio-temporal details. Second, the number of identifiable action categories is huge. This means that the same action may have different interpretations under different objects and scene contexts such as the actions \textit{drink} and \textit{eat}. Finally, phenomena such as occlusions, cluttered backgrounds, cast shadows, varying illumination conditions and viewpoint changes can all modify and influence the way the actions are perceived. Thus the sensory input and the modeling of human actions that are dynamic, ambiguous and interact with other objects are the most difficult aspects of action recognition tasks.

\begin{figure}
\includegraphics[width=0.9\textwidth]{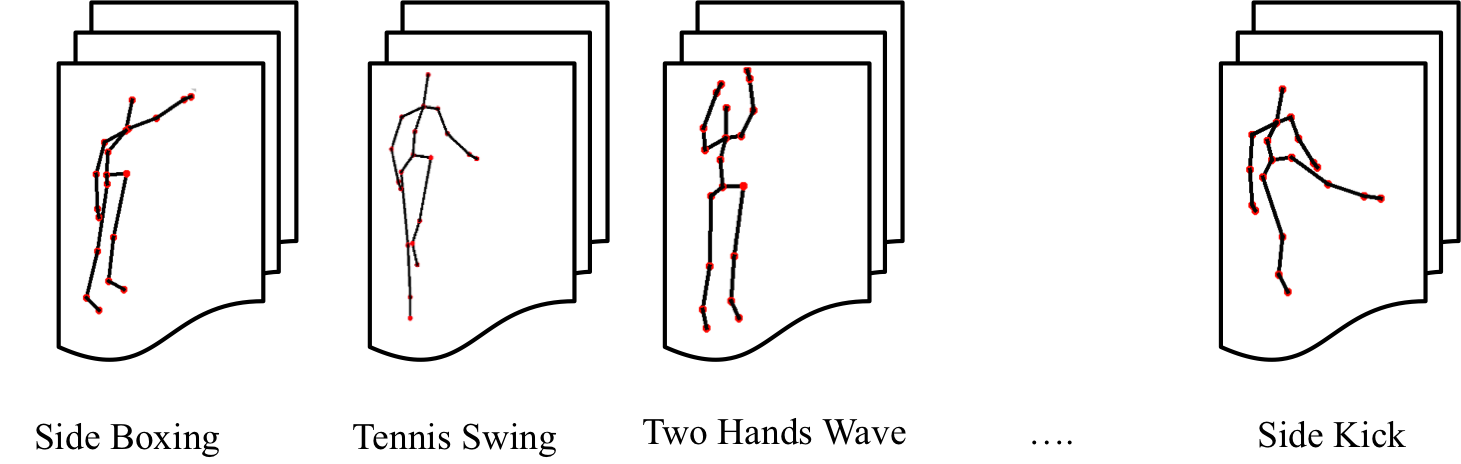}
\caption{The input space of the actions containing consecutive 3D postures of human skeleton joints.\label{fig:InputData}}
\end{figure}

With the introduction of Microsoft Kinect, a rough skeleton of the actor performing the action can be easily obtained. There are a number of benchmark datasets of actions such as MSR datasets (\cite{MSR}) that provide researchers with different types of input space such as 3D skeleton information (see Fig.~\ref{fig:InputData}). In my research, I made several experiments using MSR datasets of actions as the input to my system. I also utilized the Microsoft Kinect to collect new input datasets. The 3D positions of skeletal viewpoints were estimated from RGB and depth images with the aid of the software libraries OpenNI and NITE to read the sensors and extract the joint positions of detected human subjects in 3D space. 

The main reason I highlighted the role of input data in the action recognition task is that the techniques for extracting the action data such as 3D skeleton information, deal with the action detection problem, that is, the problem of detecting the moving figure. This problem must be solved before the action recognition can be initiated. Thus it is important to take into consideration that the action detector is strictly connected to the action recognizer and it has great influence on the action recognition performance. However, the action detection problem is not addressed in this thesis and the main goal of this study is to propose an approach for human action recognition by utilizing a 3D skeleton detector.

\section{Action recognition in robotics for human-robot interaction}
In this section, I will briefly present some of the previous research studies of action recognition in robotic platforms. Most studies aim at recognizing actions represented by result verbs. For example, in \cite{Lallee}, a new framework for embodied language and action comprehension is proposed. The action recognition system of their framework is actually a teleological representation that uses goal-based reasoning. This study aims at presenting the advantages of a hybrid teleological approach for action-language interaction both in theory and through implementation results from real experiments with an iCub robot. The experiments are based on a set of goal-directed actions such as \textit{take}, \textit{cover} and \textit{give}. The robot uses spoken language and visual perception as input and generates spoken language as output. 

One of the central functions of language is to coordinate cooperative activities (\cite{Tomasello}). Therefore the embodied language comprehension framework proposed by \cite{Lallee} is designed to connect language constructions to the actions. Their framework uses the sub-components of the actions to generate relations between the initial enabling states and the final resulting states, so-called state-action-state triples, which refer to the world states before acting, during acting and after acting. Their method makes it possible to use grammatical categories including causal connectives (e.g., because, if-then) in order to enrich the set of state-action-states that are learned. To create such a link between language and action, a neurally inspired system is constructed. The system first develops an action recognition system that extracts simple perceptual primitives from the visual scene, including contact or collision, and then composes these primitives into templates for recognizing actions like \textit{give}, \textit{take}, \textit{touch} and \textit{push}.

In the experiments by \cite{Lallee}, an iCub robot first learns from a human performing physical actions with a set of visible objects in its field of view, such as covering (and uncovering) one object with another, putting one object next to another, and briefly touching one object with another. If the robot has not seen an action before, it asks the human to describe it. The iCub learns the action description (e.g., object-1 covered object-2) and generalizes this knowledge to examples of the same action performed on different objects. The robot can also learn the causal relation between an action and the resulting change of state, for example object-1 covers object-2 and so object-2 disappears but it still exists beneath the object-1. In this scenario, actions are performed that cause changes in the final states, in terms of appearance and disappearance of the objects and so the robot should detect these changes and determine their cause. If the robot based on the knowledge gained from past experiences does not know the cause then it asks the human for clarification about the cause. 

Based on the experiment described here, the robot is able to learn for each action what is the enabling state of the world, which must hold for that action to be possible, and what is the resulting state that holds once the action has been performed. As an example if you want to take object-1 then object-1 must be visible. Therefore the state object-1 visible enables the action take object-1 and is the enabling state. On the other hand if you cover object-1 with object-2 then object-1 is invisible. Therefore the state object-1 invisible is the resulting state. This is analogous to the humans that tend to represent actions in terms of goals states that result from the performance of an action. Neurophysiological evidence of such a goal-specific encoding of actions has been collected from monkeys (\cite{Shariatpanahi}). The evidence suggests that the same action (grasping) can be encoded in different manners according to different goals (grasping for eating/grasping for placing).

An important aspect of this area of research is that the meanings of the linguistic expression are derived directly from sensory-motor experience, following embodied language processing theories. Embodied theories hold that mental simulations of the observed action are sufficient to interpret actions, while teleological theories hold that this is not sufficient. According to these theories, a generative, rationality-based inferential process is also at work in action understanding. Integrating insights from both motor-rich (simulation, embodiment) and motor-poor (teleological) theories of action comprehension is attractive as they provide different angles of the same problem.

One of the limitations of a perceptually based system is when an action causes an object to be occluded, for example when the object is moved behind another object (\cite{Lallee}). The visual disappearance of the object is totally different from the case when the object physically disappears, yet both result in a visual disappearance. The ability to keep track of objects when they are hidden during the action performs, so-called object constancy, is one of the signatures of core object cognition, which claims that human cognition is built around a limited set of \textit{core systems} for representing objects, actions, number and space (see \cite{Spelke1} and \cite{Spelke2}. 

Kalkan has developed another robotic system that can create concepts represented by verbs in language (\cite{Kalkan}). They build on the assumption that verbs tend to refer to the generation of a specific type of effect (result) in the world rather than a specific type of action. In this study the robot tries to form the concepts through interaction with the world while human uses them in order to achieve easier communication with the robot. Concepts are crucial for recognizing instances of objects. It means that to decide whether an entity is a dog, we need to have a concept of dogs. Since action concepts are analogous to objects, the same principle can be applied in the case of action recognition.

There are different views on the formation of concepts: (1) The classical rule-based view considers categories to have strict boundaries and assumes that membership of a category is based on satisfying the common properties as necessary and sufficient rules. For example if the color of the exemplar is YELLOW and the appearance of it is LONG then it is categorized as banana; (2) the prototype-based view (\cite{Rosch}) assumes no tight boundaries between categories but a \textit{prototype} (the best representing the category) that is used to judge the memberships of other items; and (3) The exemplar-based view determined by the exemplars of the categories stored in the memory (\cite{Nosofsky1}) in which any item is classified as a member of a category if is sufficiently similar to one of the stored exemplars of the category. (4) The theory of conceptual spaces where concepts are represented as convex regions in geometrically structured spaces (\cite{Gardenfors4}). This theory will be presented in greater detail later.

It has been argued that the classical view is not used in human categorization, but the evidence is unclear as to whether humans use prototypes or exemplars for generating concepts (\cite{Minda}, \cite{Nosofsky2} and \cite{Gardenfors4}). It might be that we use different types of representations and create hybrid representations for different categorization tasks (\cite{Rosseel}).

Affordances are inherent \textit{values} and \textit{meanings} of things in the environment that can be perceived and linked to the action possibilities offered to the organisms (\cite{Gibson2}). For example a \textit{chair} affords sit-ability to a human whereas it also provides hide-ability and claw-ability to a cat.

In the experiments presented in \cite{Kalkan}, the robot has been equipped with a set of behaviors (such as push left, push right, push forward, pull, top-grasp and side-grasp) in an environment with a set of objects of varying sizes and shapes, and the robot interacts with the objects to discover what they afford. There are different interactions between the robot and the objects including several interactions with the objects placed at different positions and in different orientations. Three types of features, including surface features, spatial features and object presence, are extracted from the objects before the execution of a behavior (initial features) and after the behavior (final features).

Then the difference between the final and initial features is used as the effect features. For example, if the robot applies a push-right behavior on an object, leading to a displacement towards the right, the user verbally provides moved right to the robot. The effect set is generated using a set of verbs: no-effect, moved-left, moved-right, moved-forward, pulled, grasped, knocked, and disappeared. The robot categorizes the effects of its behaviors and represents them with prototypes. They represent an effect prototype using labels ‘+’, ‘-’, ‘0’, ‘*’, corresponding to increase, decrease, no-change and unpredictable-change in the feature element, respectively. As an example of no-effect, all feature elements are represented by label ‘0’. They proposed that the effect prototypes correspond to the verb concepts. However, since \cite{Kalkan} only consider the effects of actions, the robot only learns result verbs.

The experiments presented in \cite{Stramandinoli} were intended to show that the meaning of higher-order concepts is obtained from the combination of basic sensorimotor concepts. Their network receives the action primitive words, which form the linguistic description of the higher-order word. For example in description \textit{Give is Grasp and Push and Release}, \textit{Give} is the higher-order word and \textit{Grasp}, \textit{Push} and \textit{Release} are the action primitive words. In the next step, the motor outputs corresponding to the action primitive words are computed by the network and stored one after another by keeping the corresponding temporal sequence of motor activations based on the linguistic description. As an example the sequence \textit{Grasp and Push} is different from the sequence \textit{Push and Grasp}. Finally, the network receives as input the unknown higher-order word and as target outputs the sequence of motor outputs calculated during the previous activation phase. A back-propagation technique is utilized to the new-formed training set.

The basic grounding words utilized by \cite{Stramandinoli} are \textit{Push}, \textit{Pull}, \textit{Grasp}, \textit{Release}, \textit{Smile}, \textit{Frown} and \textit{Neutral}. The first Higher-order Grounding (HG1) are: \textit{Give is Grasp and Push and Release}, \textit{Receive is Push, Grasp and Pull} and \textit{Pick is Grasp and Pull and Release}. The second Higher-order Grounding (HG2) are: \textit{Accept is Receive and Smile}, \textit{Reject is Give and Frown} and \textit{Keep is Pick and Neutral}. 

Among other systems for the action recognition in robotics is the computational architecture called HAMMER (Hierarchical Attentive Multiple Models for Execution and Recognition) proposed by \cite{Yiannis-Bassam1}. In this system, the motor control systems of the robot are organized in a hierarchically distributed manner to be used in first competitive selecting and executing an action and second perceiving an action when performed by a demonstrator. The HAMMER provides a top-down control of attention during action perception.

A content-based control of goal directed attention during action perception is also proposed by \cite{Yiannis-Bassam2}. The attended focus to the relevant body parts of the human performer is used as the input resources and applied to solve the action perception problem. Moreover the attention mechanism is augmented with another component that considers the content of the hypotheses requests to the content reliability, utility and cost. 

The proposed method in \cite{Lallee} uses the language for action recognition through verbal interactions of the human agent with the robot in which the robot will comprehend the actions from their cause and effect relations. \cite{Kalkan} uses the concepts and the similarities of the action concepts with the objects for action recognition problem. \cite{Stramandinoli} used the language for building the higher-order concepts from the combination of the basic sensory motor concepts. These approaches utilized the language (\cite{Lallee}), the concept (\cite{Kalkan}) or both of them (\cite{Stramandinoli}) for action recognition. They considered a primary assumption that the verbs tend to refer to the generation of specific types of effect in the world rather than specific type of action. 

In fact their assumption is valid only in regard to the result verbs and not the manner verbs like \text{wave}. All of the approaches mentioned earlier in this section using a human-robot interaction platform performed the experiments with the actions that result in a change in the world, such as moving an object, and thus they didn't consider a wide rage of other actions with no resulting effect in the world such as point, wave, bend, walk, run, and nod. 

The action recognition architecture proposed in this study has dealt with both groups of actions. The studies of actions, which express a manner during their performance (like waving) with no resulting change in the world are presented in papers by \cite{Gharaee3, Gharaee5} and the studies with the actions having a resulting effect in the world are presented by \cite{Gharaee2, Gharaee4} . Later in the following chapter I will describe the uniqueness and advantages of my proposed architectures in more detail.

\chapter{Proposed action recognition method}
\section{Introduction}
In this chapter I will describe the neural network algorithms used to implement the action recognition architectures. These algorithms are inspired by the neural mechanisms of the brain involved in cortical representations and reorganizations. Therefore I will start with a section explaining briefly the characteristics of the neurological functions of the cortex, in particular the topological representations of the peripheral input space. It is important to mention that for the aims of this thesis I have utilized the input data represented in the visual system. In fact among all modalities visual system plays significant role for representing and recognizing the actions, while the neural network techniques presented in this research are also applicable for other types of input space represented by other modalities such as the auditory input.

In the next sections I will elaborate how the mathematical components of the neural network algorithms implement a simple but useful model mimicking these cortical mechanisms. The main neural network components of the action recognition architecture implemented for the aims of this research are self-organizing maps and growing grids neural networks.

\section{Cortical representations of the peripheral input space}

Studies on the human nervous system show that peripheral input reaches the cortex via the thalamus from thalamocortical axons arising from appropriate thalamic nuclei. The cortex receives and processes peripheral input data in a parallel way, which results in computational parallelism (\cite{Obermayer}). The sensory cortex is made up of six layers. It seems that there is a preferred vertical flow of information from peripheral input space to the sensory cortex through the cortical layers. This path starts from Layer IV and goes to Layer II/III, followed by Layer V and Layer VI (\cite{Bolz}).

In addition to the vertical flow of information from peripheral input space to the cortex, there is a horizontal inter-connectivity between different cortical regions. These lateral interactions integrate the information from the neighboring regions of the cortical map (\cite{Buonomano}) and from the specific or distal cortical zones. The horizontal inter-connectivity seems to have a major role in cortical map reorganizations. Developing novel receptive fields and other emergent response properties after peripheral input influences in the cortical regions is of great relevance to connections between neighboring cortical areas.

An important feature of the sensory cortical areas of touch, vision and hearing is that they represent their corresponding sensory epithelial surfaces in a topographic manner. The topographic mapping of the peripheral sensory input space in the somatosensory cortex reveals itself when the neighboring cortical areas respond to the neighboring skin sites. As an example, neighboring cortical regions represent the peripheral input space from the adjacent fingers (\cite{Gazzaniga}).

The cortical representations in adult animals are not static but are continuously modified throughout life. This means that plastic changes occur in these cortical regions as a result of the influences of peripheral input. The cortex can allocate areas to represent the particular input space that is mostly used. This results in plasticity that occurs at two levels – first at the synaptic level and then in higher neuronal organizations. The synaptic plasticity (Hebbian plasticity (\cite{Hebb})) refers to an increase in the synaptic strength between neurons that fire together. At a higher level of neural organization, the Hebbian-based learning rules refer to the condition in which the peripheral inputs that fire in close temporal proximity are more likely to represent the neighboring areas of the peripheral sensory cortex.

An example of this cortical plasticity in adult animals is the results of the digit amputation in adult monkeys described in (Merzenich et al (1984)). Around two to eight months after the amputation most of the cortex area that responds to the amputated digit(s) in control animals now respond to the adjacent digits or the subjacent palm in the amputated animals. This shows that there is an expansion of cortical representation for parts of the input space that are used most (non-amputated areas adjacent to the amputated ones).

\section{Artificial neural networks inspired by the nervous system} 
It is important that an artificial architecture that has been developed to simulate cognitive process is biologically plausible. Algorithms that resemble the biological systems, such as the nervous system in humans and animals, help us to better understand these biological mechanisms. Only then will it be possible to implement optimal experiments that function similar to the biological systems.      

Based on what has been mentioned about the cortical regions of the brain, the self-organizing maps (SOM) proposed by Kohonen (1988) can be shown to possess some of the features that are essential for the architecture of the brain. Among these features are the layered and topographic organization of the neurons, lateral interactions, Hebb-like synaptic plasticity, and the capability of unsupervised learning. Another important feature of the self-organizing maps is their ability to generate low-dimensional representations of high-dimensional input spaces.

In the following sections I will describe the two different models of self-organizing neural networks in greater detail. I will elaborate the mechanisms used by these networks in order to develop a model that possesses all of the features mentioned earlier.

\subsubsection{Kohonen feature map}\label{SOM}

\begin{figure}
\includegraphics[width=0.7\textwidth]{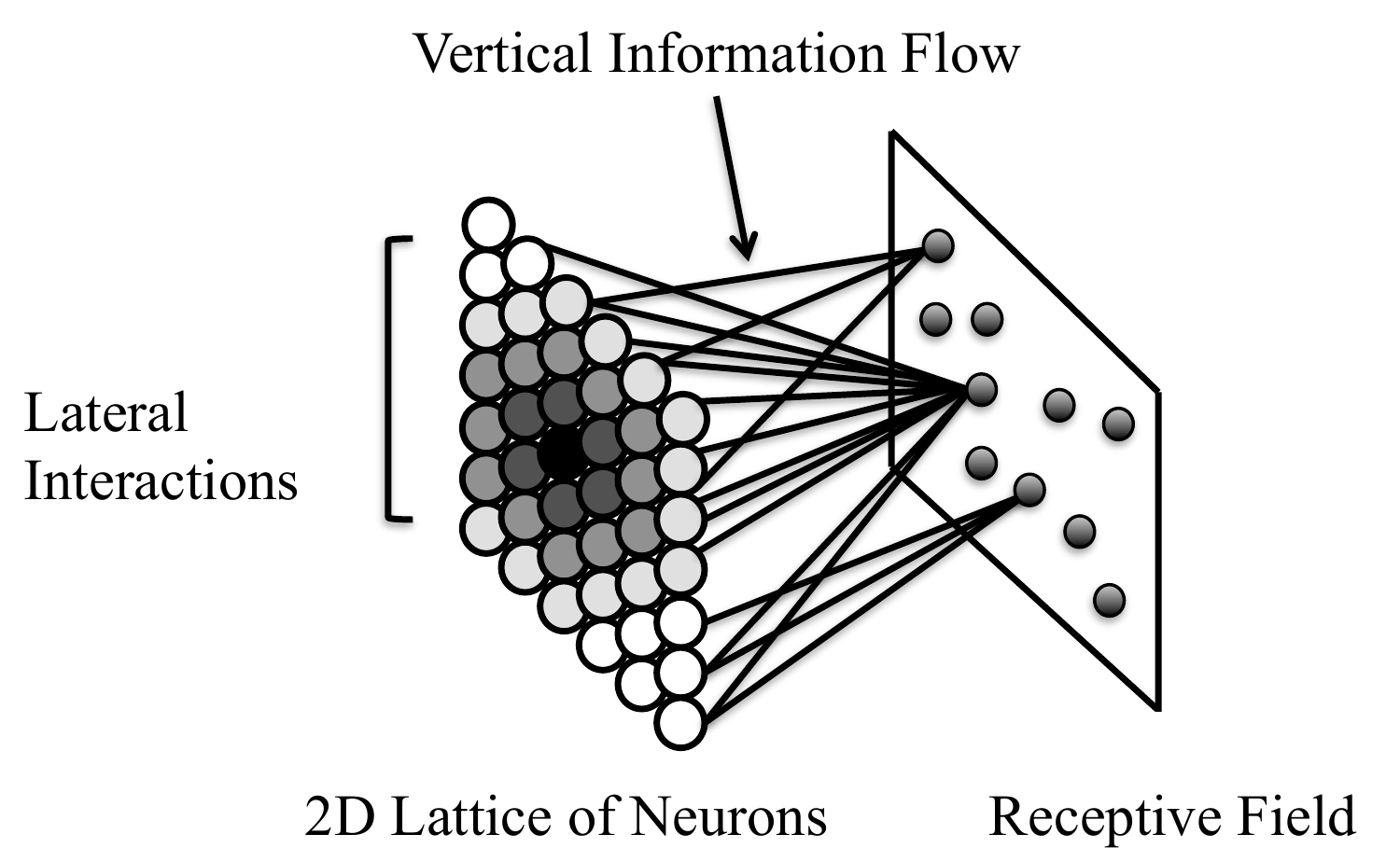}
\caption{The self-organizing feature map (SOM). The SOM consists of a 2D lattice of neurons with fixed size represented by the number of rows and columns. There is a vertical flow of information through the connections to the input space and a horizontal inter-connectivity simulated by a neighborhood function such as a Gaussian function. As shown in the 2D lattice of neurons by moving to the center of the Gaussian function the activity gets stronger (darker neurons). Thus the neurons that are further distant than the central neuron (winner) receive less activation. \label{fig:SOM}}
\end{figure}

A two-dimensional self-organizing map consists of $2D$ lattice with a fixed number of neurons and a fixed topology represented by the number of rows and columns (see Fig.~\ref{fig:SOM}). Each one of the neurons of the map is fully connected to each receptive field of the input layer. This resembles the computational parallelism in the brain (\cite{Obermayer}). Each neuron has a specific topological position (determined by the number of a row and column and represented as \textit{x} and \textit{y} coordinate of the lattice) and contains a vector of weights of the same dimension as the input vectors. If each input vector has \textit{N} dimensions then each neuron will also contain a corresponding weight vector of the same \textit{N} size.

The horizontal inter-connectivity (lateral interaction as shown in Fig.~\ref{fig:SOM}) between the neurons of the map is implemented by utilizing a neighborhood function such as a Gaussian function centered at the neuron with the largest response to an input signal (winner neuron). For each input signal the winner, which is the neuron of the map with the nearest features to the input vector, is selected and then its weight vector together with its topological neighbor's weight vectors are updated and moved towards the input signal. By applying this mechanism, the topological neurons neighboring the winner will receive partial excitation depending on their distance to the winner and at the same time the rest of the map outside this neighboring region will be inhibited.

After several adaptation steps the network will learn the structure of the input space and nearby regions of the trained map will respond to similar input data. The result is a topographic mapping of the input space. 

The version of self-organizing maps (SOMs) implemented in my thesis applies the following mathematical equations. The map consists of an $I\times J$ grid of neurons and each neuron $n_{ij}$ is associated with a weight vector $w_{ij}\in{R}^n$ with the same dimensionality as the input vectors. All the elements of the weight vectors are initialized by real numbers randomly selected from a uniform distribution between 0 and 1. 

At time $t$ each neuron $n_{ij}$ receives an input vector $x(t)\in{R}^n$. The net input $s_{ij}(t)$ at time $t$ is calculated using the Euclidean metric represented by Eq. \ref{eq:1}. Then the activity $y_{ij}(t)$ at time $t$ is calculated by using the exponential function shown by Eq. \ref{eq:2}. The parameter $\sigma$ is the exponential factor and the role of this function is to normalize and increase the contrast between highly activated and less activated areas. The neuron $w_{n}$ with the strongest activation (winner neuron) is selected by using Eq. \ref{eq:3}. 

Then the weights $w_{ijk}$ of the neurons are adapted by Eq. \ref{eq:4} in which the term $0 \leq \alpha(t) \leq 1$ is the adaptation strength, $\alpha(t) \rightarrow 0$ when $t \rightarrow \infty$ and the neighborhood function $G_{ijw_{n}}(t) = e^{-\frac{||r_{w_{n}} - r_{ij}||}{2\sigma_{r}^2(t)}}$ is a Gaussian function decreasing with time. The $r_{w_{n}}$ and $r_{ij}$ of the Gaussian function are the location vectors of neurons $w_{n}$ and $n_{ij}$ respectively and $\sigma_{r}$ sets the neighborhood radius, which covers the whole map at the beginning and decays with time.

\begin{equation}\label{eq:1}
s_{ij}(t)=||x(t) - w_{ij}(t)||
\end{equation}

\begin{equation}\label{eq:2}
y_{ij}(t)=e^{\frac{-s_{ij}(t)}{\sigma}}
\end{equation}

\begin{equation}\label{eq:3}
w_{n}=\mathrm {arg} \mathrm{ max}_{ij}y_{ij}(t)
\end{equation}

\begin{equation}\label{eq:4}
w_{ijk}(t+1)=w_{ijk}(t)+\alpha(t)G_{ijw_{n}}(t)[x_k(t)-w_{ijk}(t)]
\end{equation}

\subsubsection{Growing grid networks}
In the Kohonen feature map described above, the network is constructed from a pre-determined number of neurons represented by a fixed number of rows and columns. A more flexible implementation of self-organizing neural networks are the growing grid networks (\cite{Fritzke1}) that I will describe in this section.

In order to make a precise representation of the topology of the input space, a priori knowledge about the space is required. Constructing this knowledge requires a comparatively demanding computational task, especially in more realistic experiments. If the algorithm applies some effective heuristics to guide the development of what the architecture represents in the input space, then a more accurate topological representation of the input space can be reached (\cite{Blackmore}). The algorithms for growing grids and the growing cell structures meet these requirements. 

The growing grid network structure proposed by Fritzke meets these requirements (see \cite{Fritzke1} and \cite{Fritzke5}). Such growing networks been applied to a classification problem (see \cite{Fritzke2}), to a combinatorial optimization problem (see \cite{Fritzke3}), to a problem of surface reconstruction (see \cite{Ivrissimtzis}) and also to the touch perception in a robotic task (see \cite{Johnsson4}).

\begin{figure}
\includegraphics[width=0.7\textwidth]{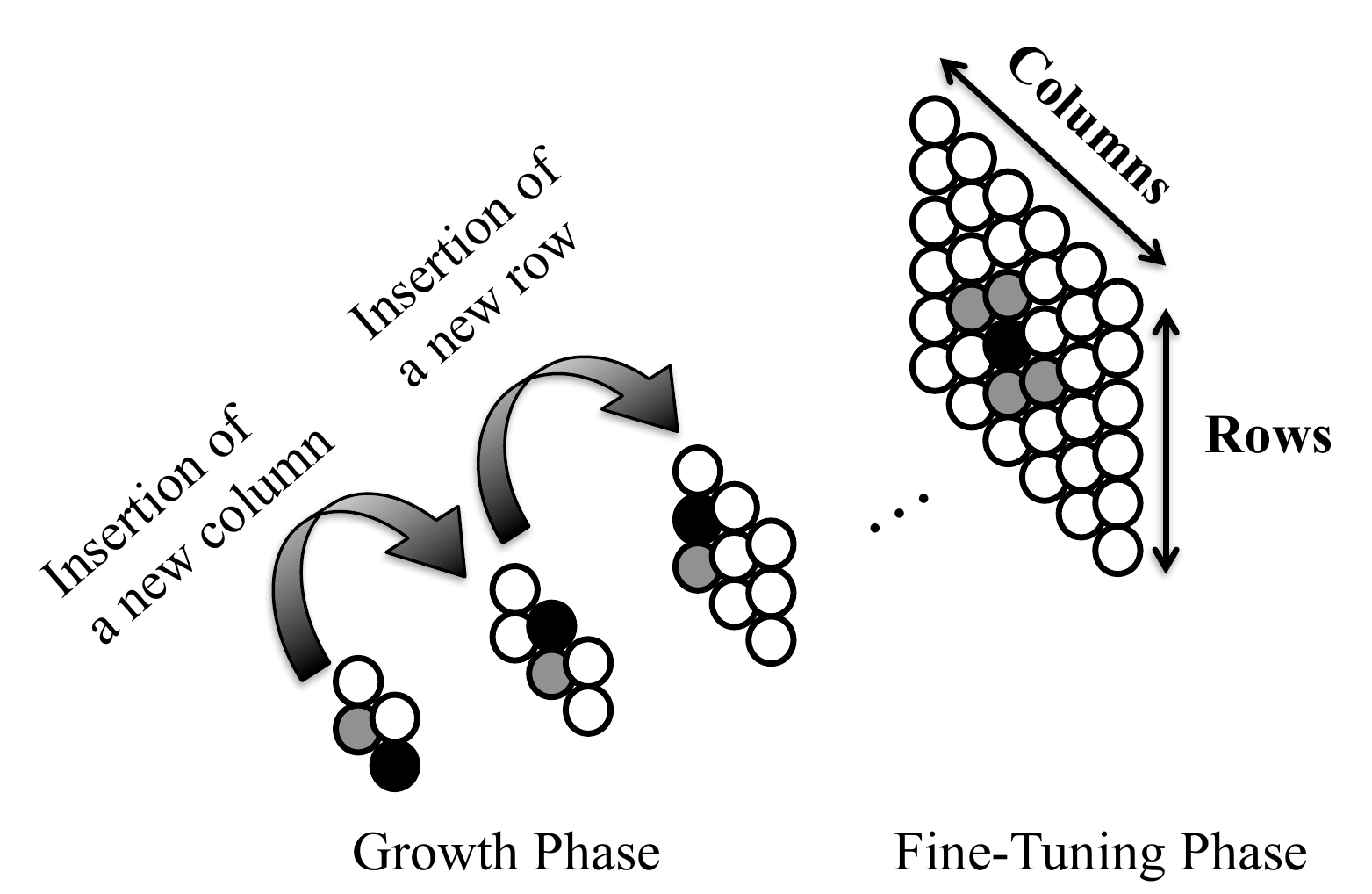}
\caption{The growing grid network. The growth phase on the left shows the expansion of the neural map by an insertion of a complete row or column of neurons in the neighborhood of the most activated areas representing the input given. The right part shows the fine-tuning phase in which the algorithm continues learning with a fixed size and a fixed topology by a decaying learning rate.   \label{fig:GG}}
\end{figure}

The growing grid is an incremental version of self-organizing feature maps with a number of neurons that increases during the learning process, but with a fixed topology. The learning in the growing grid occurs in two phases: the growth phase and the fine-tuning phase. The learning starts with the growth phase in which the map begins with a lattice of rectangular shape of size $2\times 2$ (see Fig.~\ref{fig:GG}) in which each neuron is associated with a weight vector of the same dimensionality as the input vectors. In addition to that, each neuron has a local counter value that counts the number of times at which the neuron has the largest response to the input signals during the adaptation step. 

At the start of the growth phase, the input vector $x(t)\in{R}^n$ is received by each neuron of the map. The net input $s_{ij}$ is calculated by applying the Euclidean metric (shown in Eq. \ref{eq:5}) to the input vector $x(t)$ and the corresponding neuron weight vector $w_{ij}(t)$. The activity of each neuron $y_{ij}$ is extracted by applying an exponential function (shown in the Eq. \ref{eq:6}) to the net input $s_{ij}$. The parameter $\sigma$ is the exponential factor used to normalize and increase the contrast between highly activated and less activated areas. 

The neuron $w_{c}$ that is most similar to the input vector $x(t)$ is detected by applying Eq. \ref{eq:7} and its local counter variable is then incremented by one (see Eq. \ref{eq:8}). The weight vectors $w_{ij}$ associated with $w_{c}$ and the neurons in its direct topological neighborhood are updated by using Eq. \ref{eq:9}). The learning rate $\alpha$ is a constant variable and is not a function of time for the growth phase.

\begin{equation}\label{eq:5}
s_{ij}(t)=||x(t) - w_{ij}(t)||
\end{equation}

\begin{equation}\label{eq:6}
y_{ij}(t)=e^{\frac{-s_{ij}(t)}{\sigma}}
\end{equation}

\begin{equation}\label{eq:7}
w_{c}=\mathrm {arg} \mathrm{ max}_{ij}y_{ij}(t)
\end{equation}

\begin{equation}\label{eq:8}
LC_{w_{c}}=LC_{w_{c}}+1
\end{equation} 

\begin{equation}\label{eq:9}
w_{ijk}(t+1)=w_{ijk}(t)+\alpha[x_k(t)-w_{ijk}(t)]
\end{equation}

\begin{equation}\label{eq:10}
w_{c_{1}}=\mathrm {arg} \mathrm{ max}_{ij}LC_{ij}
\end{equation}

\begin{equation}\label{eq:11}
w_{c_{2}}=\mathrm {arg} \mathrm{ max}_{pq}||w_{w_{c_{1}}}(t)-w_{pqk}(t)||
\end{equation}

After the network has received a number of input vectors, a new row or column will be inserted. The parameter $\lambda$ determines the time of a new insertion. The $\lambda$ should not be too small because the net grows too fast before being sufficiently adapted to the input space and it should not be too large because the net grows slowly due to the lack of inserted neurons to the areas of the map that represent more of the input space and require a higher resolution.

When the $\lambda$ criterion is met then the neuron $w_{c_{1}}$ with the largest local counter value is detected (see Eq. \ref{eq:10}). Among its direct topological neighbors, the neuron $w_{c_{2}}$ with the furthest distance to it is extracted (see Eq. \ref{eq:11}). If both of the neurons $w_{c_{1}}$ and $w_{c_{2}}$ are in the same row, then a new column is inserted between them and the weight vectors of the neurons in the new column are interpolated by the weight values of the neurons in the neighboring columns. Similarly, when the neurons $w_{c_{1}}$ and $w_{c_{2}}$ are in the same column, a new row is inserted between them and the weight vectors of the neurons in the new row are interpolated by the weight values of the neighboring rows. 

After the insertion is completed, the local counter value $LC$ and the $\lambda$ is reset. The growth phase continues until a performance criterion is met. Then the fine-tuning phase starts in which there will be no new insertion of rows or columns. The network will continue processing using the fixed size neural net achieved from the growth phase while the learning continues with a decaying adaptation rate. By applying several tuning steps, the network parameters will be regulated based on the input space and the learning will be accomplished.

\section{Architecture}
For the aims of this research I implemented several developments of the hierarchical SOM architecture, which is based on the self-organizing maps and the growing GG architecture, which is based on the growing grids networks. The SOMs are used for the network architectures presented by \cite{Gharaee2, Gharaee3, Gharaee4, Gharaee5, Gharaee7}. The GG networks are used for the network architecture presented by
 \cite{Gharaee6, Gharaee8}.

I will describe different components and layers of the basic SOM and GG architectures in the following sections. The basic architectures are built up of five processing layers among which there are three learning layers of neural networks. The common layers in both architectures are the preprocessing layer and the ordered vector representation layer.

\subsection{Input data and preprocessing}
The datasets utilized for running the experiments of this research are collected with the aid of Kinect sensors. The 3D joint positions of a human skeleton are extracted from the RGB image and the screen depth map of the Kinect by using the software libraries OpenNI and NITE. As a result, the body of the action performer is detected and the 3D information of its joints is extracted. There are different amounts of information available in the input space in different experiments represented by the number of detected joints. 

The preprocessing layer executes several functions including ego-centered coordinate transformation, scaling, attention filtering and extraction of dynamics. These functions are either cognitive functions such as attention and dynamic extraction inspired by human behavior in performing the same recognition tasks, or they are implemented to overcome experimental challenges such as having different viewpoints or distances towards the action performer. The functions implemented to preprocess the peripheral input data will be described in the next sections.

\subsubsection{Scaling}
\begin{figure}
\includegraphics[width=0.7\textwidth]{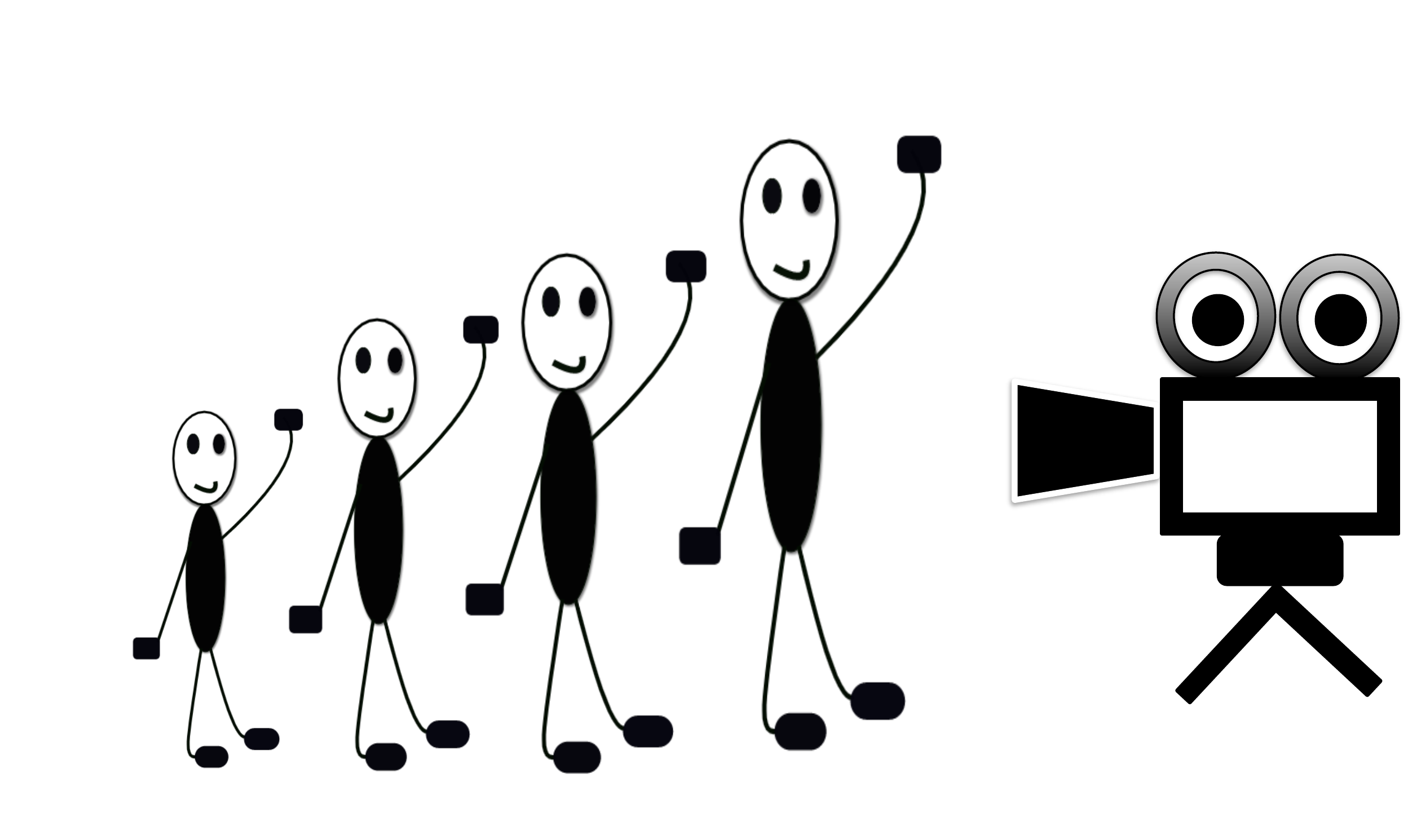}
\caption{The body of action performers have different distances to the camera which results in different scales of the bodies. All body postures are scaled into a unique size to remove this effect\label{fig:fig7}}
\end{figure}

If the action performer is placed at different distances related to the camera, then there will be a difference in the scale of the input data, which is not desirable because it could result in important spatio-temporal differences between different action sequences and thereby decrease the performance of the recognition system. Fig.~\ref{fig:fig7} shows that when the same performer moves towards the camera its spatial trajectory is recorded with different scales. 

To overcome this experimental effect the scale sizes of all Kinect skeleton postures performing different action sequences are transformed into a unique size. In this condition the input data will be invariant of varying distances to the camera. To this end let us assume that each posture of the actor's body is composed of 3D information of a certain number of skeleton joints. The adjacent joints are connected to one another by a linear rigid line called \textit{link}. This means that if each skeleton posture is represented by $N$ joints, there are in total $N-1$ links connecting these joints. I considered a standard value corresponding to each skeleton link for all posture frames of all action sequences. As an example the length of the link connecting the joint \textit{shoulder} to the joint \textit{elbow} is re-scaled to a fixed value for all action sequences by using the start joint of the link (the shoulder), its slope and the corresponding standard value.

\subsubsection{Ego-centered coordinate transformation}
\begin{figure}
\includegraphics[width=0.7\textwidth]{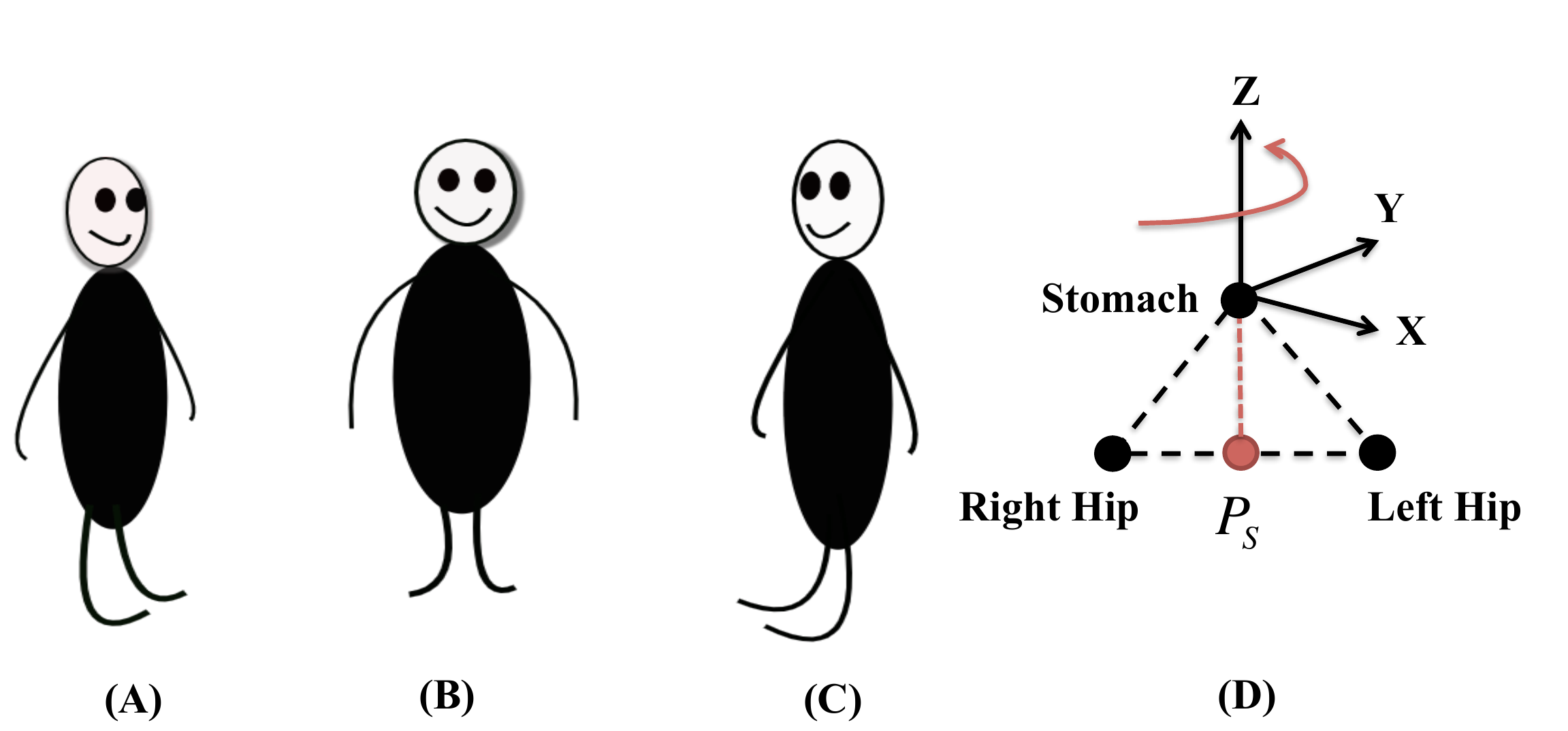}
\caption{The body of the action performers having different orientations to the camera. Turned to left (A), forward direction (B) and turned to the right (C). The coordinate system for all body postures are transformed into an ego centered coordinate system placed at the joint stomach of the body, built from the joints stomach, right hip and left hip (D)..\label{fig:fig8}}
\end{figure}

One of the main conditions that may occur in running the action recognition experiments is when there is a change in the viewpoint of the observer. This is the same as the condition when the action performers have different orientations in relation to the camera while being recorded. Fig.~\ref{fig:fig8} shows three simple examples of different orientations towards the camera when the camera is fixed in a reference frame. These examples are turning to the left, forward and turning to right, which is shown in the Fig.~\ref{fig:fig8} part(A), part(B) and part(C) respectively.

In order to overcome this condition and make the input data invariant of the orientation of the performer, I developed an ego-centered coordinate transformation. This means that all action sequences from the input data are transformed into an ego-centered coordinate system in order to be represented by one fixed coordinate system. 

The new coordinate system is called an ego-centered coordinate system because its origin is located in the joint \textit{Stomach} of the performer. The three joints \textit{Stomach}, \textit{Left Hip} and \textit{Right Hip} are utilized to build the axis of the new right-handed coordinate system as shown in the Fig.~\ref{fig:fig8} part(D). 

First the point $P_{S}$, which is the projection point of the joint \textit{Stomach} on the line connecting the joints \textit{Right Hip} and \textit{Left Hip} is calculated. The Z-axis is calculated by connecting point $P_{S}$ to the joint \textit{Stomach}. The Y-axis is directed by the vector connecting the point $P_{S}$ to the joint \textit{Left Hip}. Finally the X-axis is calculated by making a 3D cross product (vector product) of the vectors connecting the point $P_{S}$ to the joint \textit{Left Hip} and the vector connecting point $P_{S}$ to the joint \textit{Stomach}. 

Finally the origin of the new reference coordinate system is translated from the point $P_{S}$ to the joint \textit{Stomach}. All the joints 3D information is transformed into this new coordinate system.

\subsubsection{Role of Attention}
Attention is defined as focusing the mind on one of many objects or subjects that may simultaneously stimulate the mind (\cite{James}). There are many different types of attention including feature-based, object-based, temporal, spatial, bottom up and top-down attention (\cite{Gharaee1}). From the action viewpoint, attention refers to having an active observation instead of processing the entire input space, which uses a shift of attention as a means to have an active perception (\cite{Gharaee1}), for example by focusing on a moving arm instead of on the whole body. 

Attention is necessary since the limitations of time and processing power makes it necessary to apply a method based on the conscious selection of the input information and then use the selected information to perform an action such as making a decision or making a recognition (\cite{Shariatpanahi}). Moreover, the conscious selection of the input information helps to improve the efficiency of a system performance by reducing the processing of the input space to the most salient features. This is similar to visual sensors in humans in which online processing cannot process the entire input information for a space because of limitations in visual field, time, and processing power (\cite{Shariatpanahi}).  

Learning how to control attention to a perceptual subspace while learning a specific task results, first, in saving time and processing power and, second, in improving the performance of the learned task (\cite{Gharaee1}). To this end, I have implemented a manual determination of the salient joints for the set of actions in the preprocessing layer of the architecture. This resembles the condition in which the agent sets the focus of attention to the most salient parts of the body involved in performing the action and then uses this selected information to complete the task of recognizing that action. 

It seems that humans apply a similar mechanism for recognizing the actions performed by others. As an example, when we observe a person waving to greet somebody by lifting the arm, our attention is mainly focused on the arm involved in performing the action \textit{wave} and in ignoring other parts of the body such as the legs.

\subsubsection{Applying orders of dynamics}
Next, I turn to how the action categories can be modeled. Here I built on the theory of \textit{conceptual spaces}. As G\"ardenfors defined it, a conceptual space is built up from geometrical representations based on a number of quality dimensions (see \cite{Gardenfors4}). Examples of the quality dimensions are temperature, weight, brightness, pitch and the spatial dimensions height, width and depth. We perceive the spatial dimensions of height, width and depth as well as brightness by our visual sensory system, pitch by the auditory system, temperature by thermal sensors and weight by the kinesthetic sensors. There are also many quality dimensions that are of an abstract character.

What G\"ardenfors means by geometrical representation can be clarified by a domain involving our \textit{color perception}. Our cognitive representation of the color can be described by three dimensions \textit{hue}, \textit{saturation} and \textit{brightness}. The first dimension \textit{hue} starts from \textit{red} via \textit{yellow}, goes to \textit{green} and \textit{blue} and back again to \textit{red}, which generates a circle. The second dimension \textit{saturation} ranges from \textit{grey} with zero intensity and increases to greater intensities, which is isomorphic to an interval of the real line. The last dimension \textit{brightness} ranges from \textit{white} to \textit{black} and so resembles a linear dimension with end points.    

\begin{figure}
\includegraphics[width=0.55\textwidth]{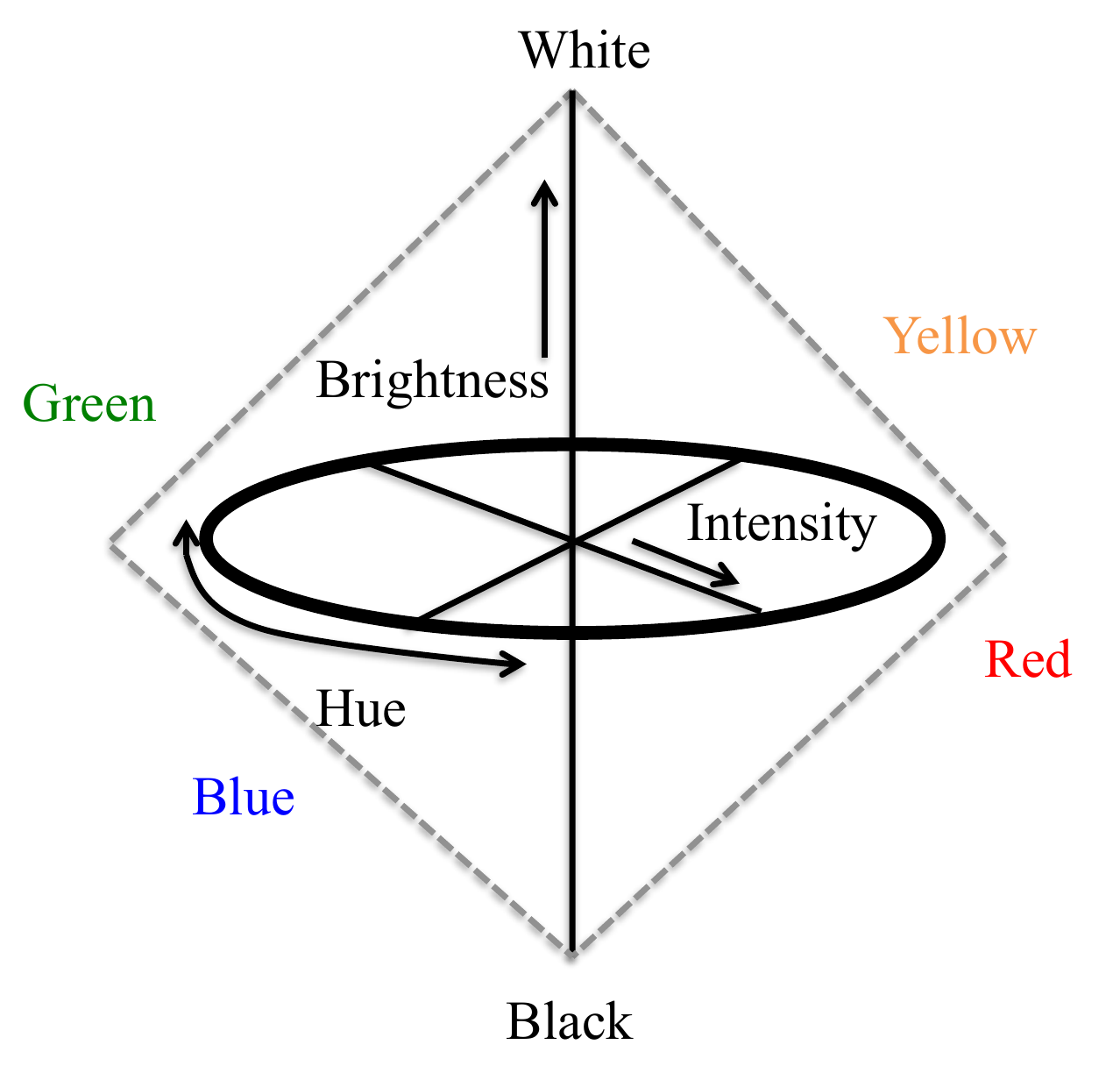}
\caption{The color spindle\label{fig:fig9}}
\end{figure}

The color domain as a subspace of our perceptual conceptual space is often illustrated by a \textit{color spindle}, which is shown in Fig.~\ref{fig:fig9}. There, \textit{brightness} is shown on the vertical axis, \textit{saturation} is shown as a distance from the center of the spindle and finally \textit{hue} is represented by position alongside the perimeter of the central circle.

A conceptual space can be defined as a collection of quality dimensions. The dimensions can be correlated in various ways such as the ripeness of the fruit being correlated with its color. It is not possible to provide a complete list of the quality dimensions involved in the conceptual spaces of human cognition. We know that some of the dimensions are innate and partly hardwired in the nervous system such as color, pitch and space. Other dimensions are learned, which results in the expansion of the conceptual space by new dimensions during our development. There are other types of dimensions such as culturally dependent dimensions and dimensions introduced by science.

The question regarding the action categories is what dimensions they are based on. According to the conceptual spaces theory of G\"ardenfors, we can represent \textit{actions} by the \textit{force patterns} that generate them. The \textit{forces} represented by the brain are psychological constructs and not necessarily the scientific dimension introduced by Newton. According to this theory, the information that our senses such as vision receive by observing the movements of an object or an individual is used by our brain to extract the underlying force patterns in an automatic way. According to Runesson’s KSD principle, this information is available in our optic array in the same way as the information about other properties like the color or shape of an object.

Thus by adding the \textit{force} dimensions to a conceptual space we gain a basic tool to analyze the dynamic properties of actions and other types of motion. To represent the action by the force patterns generated the action, I calculated and applied the orders of dynamic. An action can be represented by both the kinematic and the dynamic characteristics visible in the movements of an object or an individual. These characteristics can be determined from the movements of the joints and the skeleton parts and so can be extracted from the skeletal information.

From the 3D position of the skeleton joints I calculated the first-order dynamics, which represents the velocity of the corresponding joint, as well as the second order of dynamics, representing the acceleration of the joints. As we know from Newton’s law, the second order of dynamic representing the acceleration vector has the same direction as the force vector and the magnitude of the force vector is inversely proportional to the mass of the moving object.

Thus by applying the first and second orders of dynamic as the input, one can extend the input space to contain more information about the action. Some of this information represents the force patterns. As a result, the system will receive richer information about the dynamic and kinematic properties of the actions and may use this information for recognizing the actions. I investigated the effects of including first- and second-order dynamics presented by \cite{Gharaee5}.

\subsection{The role of different layers of the architecture}
\begin{figure}
\includegraphics[width=0.75\textwidth]{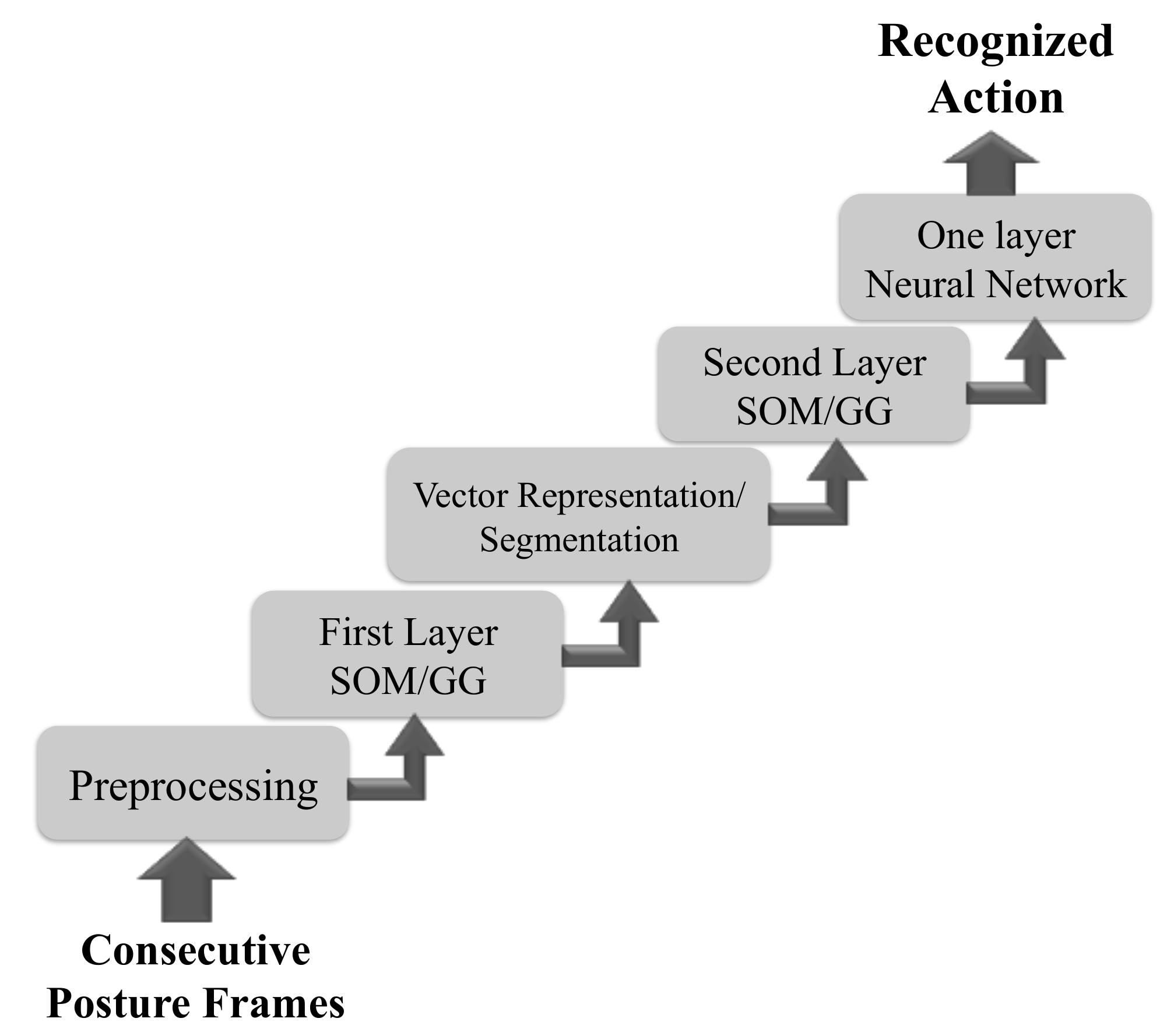}
\caption{The multi-layer action recognition architecture\label{fig:Architecture}}
\end{figure}

In this section I will describe how different layers of the hierarchical action recognition architectures are connected. As represented in Fig.~\ref{fig:Architecture}, the system is composed of five main processing components. Among them are three layers of neural networks, one layer preprocessing and one layer of ordered vector representation.

The first-layer neural network, which can be a SOM or a GG layer, receives the preprocessed input data from actions as a series of consecutive 3D posture frames. Each posture frame elicits activation in the first layer neural map and by connecting these consecutive elicited activations a pattern vector is generated for each action sample. 

Series of activations are received by the ordered vector representation layer to create time invariant pattern vectors of key activations. The consecutive activations will be represented as a unique action pattern. Then the generated action patterns will be sent to the second layer neural network.

The second layer neural network, which again is either a SOM or a GG layer, receives the action pattern vectors and categorizes them into action categories. Based on how differently one action is performed there may be several sub-clusters representing the same action, which makes the classification task more complicated. 

Finally, the third layer of the architecture, which is a one-layer supervised neural network, labels the clusters/sub-clusters shaped by the second layer neural map with the action labels. Any new test action sample that is received by the architecture with trained weights will be assigned an action label, if it elicits the area of the neural map in the proximity of the corresponding action cluster/sub-clusters.

I tried to propose an action recognition architecture which has several features and among them are its biological inspiration from the living organism, particularly the human, in performing the same action recognition task. Using the three layers of neural network in hierarchical action recognition architecture introduces a semi-supervised learning model (\cite{Ding}), which resembles the human learning process in which the training samples are often obtained successively. In this way the observations arrive in sequence and the corresponding labels are presented very sporadically. Moreover the self-organizing maps and growing grid networks have other important features such as topographic organization of the neurons, lateral interactions, Hebb-like synaptic plasticity, and the capability of unsupervised learning. 

Another important feature of my proposed architectures is their ability to generate low-dimensional representations of high-dimensional input space. The system receives spatio-temporal input data of action sequences from large number of posture frames and produce action pattern vectors, which represent the spatial as well as the temporal information of the action sequences.

Most of the methods for action recognition utilize the pre-segmented and labeled datasets of actions (see \cite{Li}, \cite{Liu-Mengyuan}, \cite{Wang2}, \cite{Yang-Xiaodong1}, \cite{Shotton} and \cite{Oreifej}), while online recognition of actions is crucial in real-time experiments with unsegmented sequences of actions. In paper by \cite{Gharaee7}, I presented the developed online action recognition architecture utilized in performing experiments on unsegmented sequences of actions.

The deep learning approaches for action recognition such as the CNN-based and RNN-based systems have certain features and among them is that these neural networks are not biologically inspired. Moreover working with deep learning techniques requires to train the large number of deep network's parameters with a big input data. While in many tasks such as medical tasks we don't have access to a big amount of input data or it is too expensive to produce it.

Many of the proposed approaches for the action recognition in a robotic platform utilize as their input space only the result actions, the actions result in a change in the surrounding environment of the observer such as a move in an object (see \cite{Lallee}, \cite{Kalkan} and \cite{Stramandinoli}). The action space is wide and also includes the actions merely expressing a manner without having a change in any entity existing in the surrounding environment of the observer such as the actions wave, jogging and point. The proposed action recognition architecture of this thesis deals with both groups of actions. The actions without involvement of another entity are recognized in the experiments presented in papers \cite{Gharaee3, Gharaee5} and the result actions are considered as the input of the system presented in papers \cite{Gharaee2, Gharaee4}.

In the following chapter I will explain how the architectures are developed to implement different experiments in online real time mode with unsegmented action sequences. The more detailed descriptions of different experiments are available for the reader in my research articles.

\chapter{Segmenting actions}
\section{Introduction}
One of the main purposes of developing an action recognition system is that is can be used in several kinds of applications such as video surveillance, health care system or human-robot interaction. However, for such applications the system must function online in real time. This means that the system has to be capable of dealing with unsegmented and unlabeled input sequences of actions. In \cite{Gharaee7}, I presented the results of experiments I have performed for online recognition of unsegmented actions, using an extended hierarchical architecture. 

The segmentation problem arises in case of receiving a continuous stream input data of actions. This means that it is not announced to the system when an action starts or when it ends. Consequently, the system has to determine these conditions on its own. In this chapter, I will investigate how the system can be made capable of recognizing actions from an unsegmented input data in realistic experiments. I will also compare this to what is known as human event segmentation tasks. 

Our daily activities are goal-driven and these goals are indirectly utilized by our perceptual mechanisms. The goals provide major causal constraints on our behavior as well as on how our actions are segmented. One possibility for someone trying to comprehend a series of events is to monitor the goals and sub-goals of the actors in the events and to segment activity into events corresponding to the goal units. This idea individuates actions in terms of the actor's intentions, while the goals covary with other information in human activities that change over time.

The goals of an actor are normally internal features and not visible to the observer, while the features such as the actor's movements, changes in the objects, people and locations are directly visible to the observer and they can reliably be related to the internal goals. Thus the correlations among these features can lead and govern the event comprehension where the observer tracks the physical movements and visible features until a change is detected and sets the event boundary at the point where a greatest change is identified.    
 
This proposal of utilizing changes in perceptual features for event segmentation faces two main problems. The first problem comes with what is considered as a change in the stimuli features; in a case of tracking, for example, the position of an object over time when a change occurs whenever the object moves. While if we take the velocity as the stimuli feature, then a constant motion is considered as no change. The second problem is when the change theory does not explain how the segmentation interacts with other components of perception and thought such as attention, memory and planning.
   
In the following section, I will first describe a theory regarding event perception in humans and then I will explain how this might lead us to explore possible solutions in dealing with the same problem in artificial systems.

\section{Event segmentation theory} 
In this section, I will describe the Event Segmentation Theory (EST) suggested by \cite{Radvansky}. When we are observing daily activities, we constantly face lack of information, for example concerning the goals of actions. For the goals of other people, the lack of information is due to the fact that goals are invisible and so we are unable to observe them directly. But also for simple perceptual features such as the location of an object, we often lack information as a result of lack of attention, sensory limitations or occlusion. To deal with this condition, perception is augmented with memory systems to keep the representations of current activities that are not actively perceived.

Radvansky and Zacks call these representations \textit{working models} and defined three main properties for them. The first relates to their limited duration, in contrast with memory representations that maintain large quantities of information over long time intervals. The second feature of the working models is that they represent the features of the current activity that are relevant to one's current goal and task. Finally, the working models are multimodal and they integrate information of various modalities with conceptual information.  

Such working models improve the comprehension of an event by biasing the pathway from sensory inputs to predictions in order to be protected from the moment-by-moment changes in the sensory space. EST suggests that most of the time working models are isolated from the input space and just store a snapshot of the current event. However, the working models cannot be isolated from the input space forever and there is therefore a need to establish a balance between stability and flexibility.

EST suggests a trade-off between stability and flexibility, which is accomplished by tracking the proximity between the prediction of the near future and what actually is happening (extracting the \textit{prediction error}). Thus the observer constantly calculates the prediction error and when something unexpected happens so that the prediction error jumps from a low value, the observer updates the working model of the event.

Updating the working model necessitates transient opening of the inputs while there are two main input resources. The first is the current state of the sensory and perceptual world and the second is the long-term knowledge of the event categories and their structures. Thus when the gates are opened, the perceptual information interacts with knowledge representations to create a new working model of the ongoing event. As a result of this procedure the prediction error decreases and the gates are closed. 

In current artificial systems there are two main phases in almost all cases: the learning phase and the test phase. During the learning phase the knowledge of the system is built and the memory system is created to maintain information obtained over the learning period. In the test phase the knowledge representation and memory storage stops and disconnects from the sensory input and the system is assumed to utilize the stored information and built knowledge to face any test input space. In EST theory, however, the knowledge-based system partly represented by the working models does not remain isolated from the input space forever and it is updated if the prediction error jumps from a certain value. This represents that the knowledge built so far is not sufficient to face the current condition. Thus the EST theory might be applicable only for the learning phase of the current artificial systems in which the knowledge representations and the memory systems are connected to the input space and can be updated.

\section{Segmentation model}
Assume we have available continuous input data of actions, without access to a manual segmentation process, which identifies when an action starts and when it ends. In order to make online experiments on unlabeled input data of actions in an artificial action recognition system, it is then necessary to model the segmentation process. 

The segmentation models that have been presented in the literature largely depend on which sensor process is used to build the input space. When the input data of actions consists of RGB images, action segmentation is more similar to object segmentation in which the visual scene is perceived as consisting of a ground that is less salient and less differentiated while there are one or more figures that are perceived as salient components (spatial kinematics represented by the actors bodies). In contrast, in my research I used input data of actions consisting of consecutive postures representing the 3D skeleton information of the joints. 

As I have already discussed, actions have spatio-temporal characteristics. This means that there is kinematic information about the actions. The temporal information represents the time interval an action lasts as well as the temporal order of the movements that are most significant for recognizing the actions. As an example, assume that you lift up your arm and then push it forward. This represents the action \textit{throw} or \textit{punch} while if you take the reverse path – push back your arm and put it down – the action \textit{catch} is represented.  

To segment the actions, it is important to define the key postures for each sequence of an action, which contains the main movements that are critical to complete that action. The action is identified if these key postures represented by the body of the actor are observed by the system. In fact, the system is tracking the movements of the actor and is looking for these key components to detect what action is performed. As an example, for the action \textit{scratch the head} these key components are: the arm is lifted up, the hand approaches the head and after while the hand leaves the head. No matter how each one of these intermediate steps is performed, if the system detects the key component of each step the action is identified.

The technique I used for recognizing unsegmented actions in an online mode is partly similar to the EST theory. Indeed, by applying self-organizing maps, an automatic segmentation mechanism has been constructed.

As mentioned in section \ref{SOM}, the first layer of the self-organizing map architecture extracts the activations representing the posture frames of an action sequence. The connections between the first and the second layer of the architecture create the action pattern vectors by connecting unique elicited activations extracted from the first layer self-organizing map. By setting a sliding window on the action pattern vector of all consecutive actions the second layer self-organizing map receives the patterns of key activations and categorizes them.

This means that when the SOM system receives initial postures of an action performed it extracts the key activation and waits for the remaining key postures to make a certain recognition based on the learned concepts. This is partly similar to the EST theory, which claims that the event's observer predicts the near future based on his/her stored memory and knowledge and compares it with the real future. In fact in both cases based on the initial input data as well as the learned knowledge and memory the agent expects for the occurrence of specific movements and if it is not satisfied then the agent consider it as the end of an event and the beginning of a new event. 

There is delay in recognizing an action, which is due to the fact that it takes some time for the system to receive all key postures and consequently create the corresponding key activation vector. This is compatible with experiments with human observers since it takes time for us to recognize an action with certainty so we wait until we receive the necessary information concerning the key components of the action. A more detailed description of the segmentation techniques I have utilized in SOM architecture is presented by \cite{Gharaee7}.

\chapter{Experiments and results}
\section{Introduction}
In this chapter I will briefly describe the experiments that have been performed for this thesis. The details of the experiments are provided in the following articles.

In total, the articles present six main experimental setups that have been designed and implemented. The setups address different problems and angles of the action recognition tasks. The main purpose of this thesis is to develop an efficient action recognition framework in a robotic context. In other words, the action recognition architectures developed aim at making the robot capable of perceiving human actions and utilizing the perceived knowledge to build better human-robot interaction in a social environment.

It should be clear that the action recognition architectures proposed in this research can be employed in many experimental conditions that are suitable for the goals of different applications such as gaming. Therefore, I have developed a human action recognition method that can be used for a wide range of applications.

\section{Experimental results}
In the first experiments, the aim was to develop an action recognizer that deals with actions only, without any other objects being involved. The experiment, presented by \cite{Gharaee3}, was implemented in an offline mode on already segmented actions of the publicly available dataset MSR-Action3D (\cite{MSR}). In different experiments the total number of action sequences is divided into training and test sets in which the training sets contain $75\%$ to $80\%$ of action sequences and the test sets contain the rest $20\%$ to $25\%$ of the action sequences selected randomly from the whole datasets.

The test results were that $83\%$ of the actions not included in the training set were correctly categorized. For the second experiment, presented by \cite{Gharaee2}, I developed an online action recognition system using the neural modeling framework Ikaros by \cite{Balkenius} to run real-time experiments. I collected two datasets of actions with Kinect using the software libraries OpenNI and NITE to read the sensors and extract the 3D joint positions of the action performer. I first evaluated the system in the experiment with actions without object and achieved categorization test accuracy of $88.6\%$. Then I evaluated the system in experiments with actions including objects and achieved a categorization test accuracy of $94.2\%$.

In the third experiment, presented by \cite{Gharaee4}, I developed an architecture that can also handle the information about objects by utilizing an object tracker module of the Ikaros. To this end I developed a hybrid system with two parallel paths for processing information of the actor and the objects. This system could recognize what action is performed and also could identify which object the actor acted on among the four available objects in the environment. The action categorization accuracy was $91.1\%$ and the object identification accuracy was $100\%$.

In the fourth experiment, presented by \cite{Gharaee5}, I extended the architecture to be able to handle the information about the first and second orders of dynamics of the actions performed. The merged action recognition system receives the information of the joints 3D positions together with their first and second orders of dynamic representing the velocity and acceleration of the joints. Several experiments were performed on two datasets of actions that contain in total 20 actions. I compared the categorization accuracy with and without merging the dynamics information. For the first dataset I achieved an improvement in accuracy from $83\%$ to $88\%$ and in the second experiment I achieved an improvement in accuracy from $86\%$ to $90\%$ by application of the two orders of dynamics.

In the fifth experiment, presented by \cite{Gharaee7}, I developed a new version of the system that is capable of recognizing unsegmented and unlabeled human actions in an online real-time experiment. The performance of the system was evaluated in two main experiments and I achieved a categorization test accuracy of $75\%4$ and $88.74\%$ respectively. 

Finally, for the sixth experiment, presented by \cite{Gharaee6}, a new version of architecture was developed by implementing two layers of growing grid neural networks. This new architecture was tested on different datasets of actions and compared with the architecture based on self-organizing maps. The first experiments involved 10 actions and the system achieved $93.00\%$ of categorization accuracy while the SOM architecture only had a $90\%$ accuracy. The second experiment involved 20 actions and the system achieved a categorization accuracy of $71.20\%$, which is significantly superior to the SOM architecture that only had a $59.61\% $ accuracy. There was also a significant increase in the learning speed of the growing grid system compared to the SOM system.

\section{Action recognition applications and future works}
There are many different applications for artificial action recognition systems. Specifically, in robotics an action recognition system can be employed in social robotics where there is a need to interact with robots. Assistant robots in health care applications are another application area. In general in any social context where there is a need for interaction between a human agent and another artificial agent, there will be a need for action recognizer frameworks.

In summary, I have designed and developed hierarchical action recognition architectures applying self-organizing map (SOM) and growing grid (GG) neural networks equipped with different processing layers. These architectures are developed with different applications such as the abilities to recognize human actions represented by either manner or result verbs, to identify the object involved in the performance of the actions, to perform the online real-time experiments. Different abilities of the architectures have been evaluated in several experiments represented in detail in the scientific articles.  

Despite all the abilities of the proposed action recognition architectures, there are many steps to improve the system and make it applicable in running different experiments with a wide range of actions in varying experimental conditions. Among the many factors that can be developed are improvements of the type of input that provides the system with data of the actions. In particular, there is a great need for an efficient action detector that is implemented one step prior to the action recognizer. Developing the preprocessing mechanisms could be another future step. As an example developing an automatic attention mechanism that builds a salience map of the scene, and extracts the attention focus of the observer while the action is performed will generate a more robust action recognition system. 

Developing the system to be able to categorize a wide range of actions – with and without involvement of other objects, or actions performed in an interaction between two actors is another important step that should be taken in the future. Improving the method of the action key features extractor and also the action classifier should be considered as other important plans for building an efficient and robust action recognition system.

%%\blindtext
%%\blindtext
%%\blindtext
%The concepts are summarized in Table~\ref{tab:comparison}.
%% TABLE SATELLITE PARAMETERS ------------------------------
%\begin{table}
%\caption{Table caption with a CAPS word, no small caps can be used in table and figure captions because sans serif fonts don't support it.\label{tab:comparison}}
%\centering
%\begin{tabular}{lcccc}
%\toprule
%& Superman & Spiderman\\
%\midrule
%Gender & male & male\\
%Species 	& Homo Sapien 	& Human/spider\\
%Homeworld 	& Gotham City 	& Earth\\
%\midrule
%Publisher 	& DC Comics 	& Marvel Comics\\
%\bottomrule
%\end{tabular}
%\end{table}
%
%
%
%\blindtext

%\begin{thesisbox}{Paper \I}
%\textbf{Brief headline summarizing its contents}
%\\[0.3em]
%\blindtext
%\end{thesisbox}

% ===============================================================
% ====================== References:  ===========================

%% Bibliography renamed to references
\newpage

\addcontentsline{toc}{chapter}{References}
\renewcommand{\bibname}{References}
\bibliographystyle{apa}
\bibliography{ZahraReferences.bib}  % Link to bibtex file %%%%%%%%%%%%%%%%%%%

\end{document}